\documentclass{article}

\PassOptionsToPackage{numbers, compress}{natbib}
\usepackage[final]{neurips_2022}

\usepackage[utf8]{inputenc} % allow utf-8 input
\usepackage[T1]{fontenc}    % use 8-bit T1 fonts
\usepackage{hyperref}       % hyperlinks
\usepackage{url}            % simple URL typesetting
\usepackage{booktabs}       % professional-quality tables
\usepackage{amsfonts}       % blackboard math symbols
\usepackage{nicefrac}       % compact symbols for 1/2, etc.
\usepackage{microtype}      % microtypography
\usepackage{xcolor}         % colors
\usepackage{multirow}% colors
\usepackage{graphicx}
\usepackage{enumitem}
\usepackage{subfigure}
\usepackage{amssymb}
\usepackage{fdsymbol}
\usepackage{cancel}
\usepackage{xcolor}
\usepackage{wrapfig}
\usepackage{caption}
\usepackage{blindtext}
\newtheorem{myDef}{Definition}
\usepackage[ruled,linesnumbered,lined]{algorithm2e}
\newcommand\Ccancel[2][red]{\renewcommand\CancelColor{\color{#1}}\xcancel{#2}}

\title{Debiasing Graph Neural Networks via Learning Disentangled Causal Substructure}

\author{
Shaohua Fan\textsuperscript{1,2}\thanks{This work was done when the first author was a visiting student at Mila.},
Xiao Wang\textsuperscript{1},
Yanhu Mo\textsuperscript{1},
Chuan Shi\textsuperscript{1}\thanks{Corresponding authors.},
Jian Tang\textsuperscript{2,3,4 \dag}\\
\textsuperscript{1}Beijing University of Posts and Telecommunications, China\\
\textsuperscript{2} Mila - Qu\'ebec AI Institute, Canada\\
\textsuperscript{3} HEC Montr\'eal, Canada\\
\textsuperscript{4} CIFAR AI Research Chair\\
\texttt{\{fanshaohua, xiaowang, moyanhu, shichuan\}@bupt.edu.cn}, \texttt{jian.tang@hec.ca}
}

\begin{document}

\maketitle

\begin{abstract}
Most Graph Neural Networks (GNNs) predict the labels of unseen graphs by learning the correlation between the input graphs and labels. However, by presenting a graph classification investigation on the training graphs with severe bias, surprisingly, we discover that GNNs always tend to explore the spurious correlations to make decision, even if the causal correlation always exists. This implies that existing GNNs trained on such biased datasets will suffer from poor generalization capability.  By analyzing this problem in a causal view, we find that disentangling and decorrelating the causal and bias latent variables from the biased graphs are both crucial for debiasing. Inspiring by this, we propose a general disentangled GNN framework to learn the causal substructure and bias substructure, respectively. Particularly,  we design a parameterized edge mask generator to explicitly split the input graph into causal and bias subgraphs. Then two GNN modules supervised by  causal/bias-aware loss functions respectively are trained to encode causal and bias subgraphs into their corresponding representations. With the disentangled representations, we synthesize the counterfactual unbiased training samples to further decorrelate causal and bias variables. Moreover, to better benchmark the severe bias problem, we construct three new graph datasets, which have controllable bias degrees and are easier to visualize and explain. Experimental results well demonstrate that our approach achieves superior generalization performance over existing baselines. Furthermore, owing to the learned edge mask, the proposed model has appealing interpretability and transferability.\footnote{Code and data are available at: https://github.com/googlebaba/DisC.}
 
\end{abstract}

\section{Introduction}
Graph Neural Networks (GNNs) have exhibited powerful performance on graph data with various applications~\cite{kipf2016semi,velivckovic2017graph,hamilton2017inductive,fan2019metapath,fan2020one2multi}.  One major category of applications are the graph classification task, such as molecular graph property prediction~\cite{hu2020open,lee2018graph,ying2018hierarchical}, superpixel graph classification ~\cite{hendrycks2019benchmarking}, and social network category classification~\cite{zhang2018end,ying2018hierarchical}. It is well known that graph classification is usually determined by a relevant substructure, but not the whole graph structure~\cite{ying2019gnnexplainer, lucic2021cf, yuan2021explainability}. For example,  for MNIST superpixel graph classification task, the digit subgraphs are causal ($i.e.,$ deterministic) for labels~\cite{vu2020pgm}. The mutagenic property of a molecular graph depends on the functional groups ($i.e.,$ nitrogen dioxide ($\text{NO}_2$)), rather than the irrelevant patterns ($i.e.,$ carbon rings)~\cite{luo2020parameterized}. Therefore, it is a fundamental requirement for GNNs to identify causal substructures, so as to make correct prediction. 

\begin{figure}[!htbp]
\centering
\subfigure[Examples of graphs in CMNIST-75sp.]{
\includegraphics[ width=7cm]{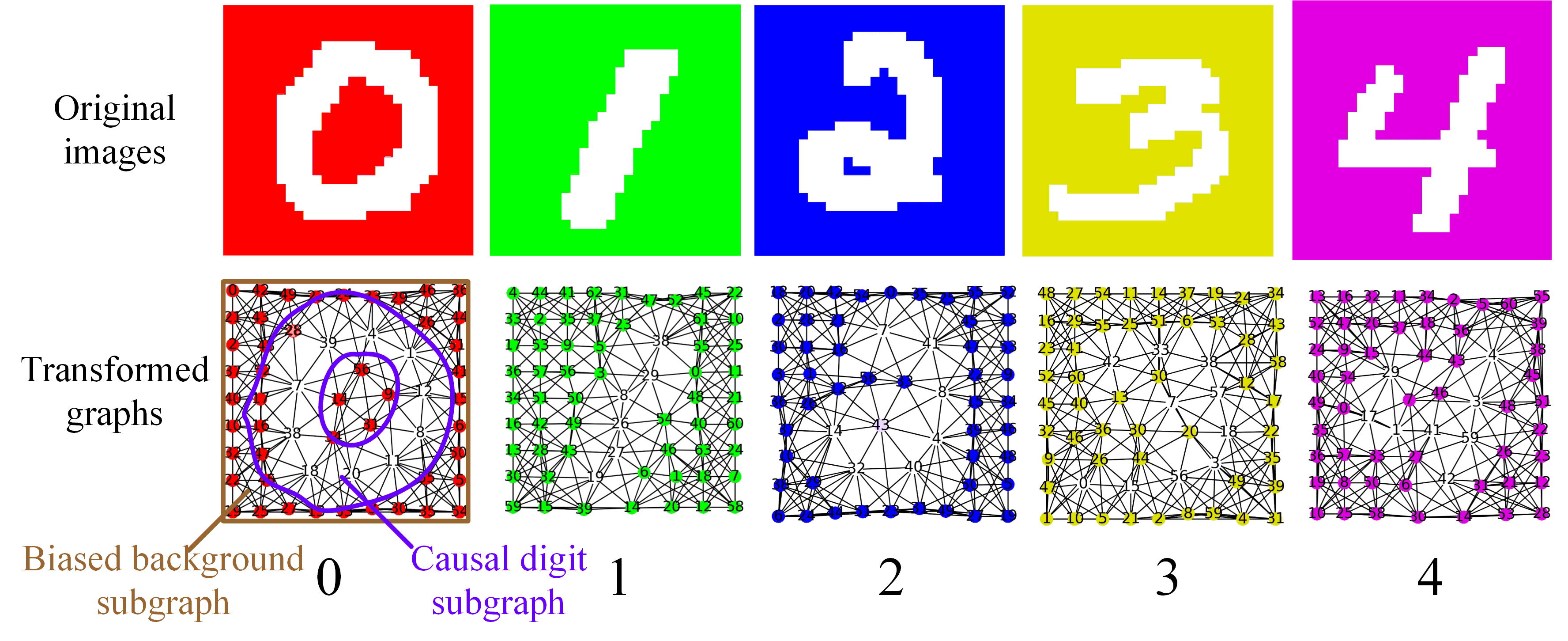}
\label{fig:illustration}
}
\subfigure[Performance of GNNs.]{
\includegraphics[ width=4.5cm]{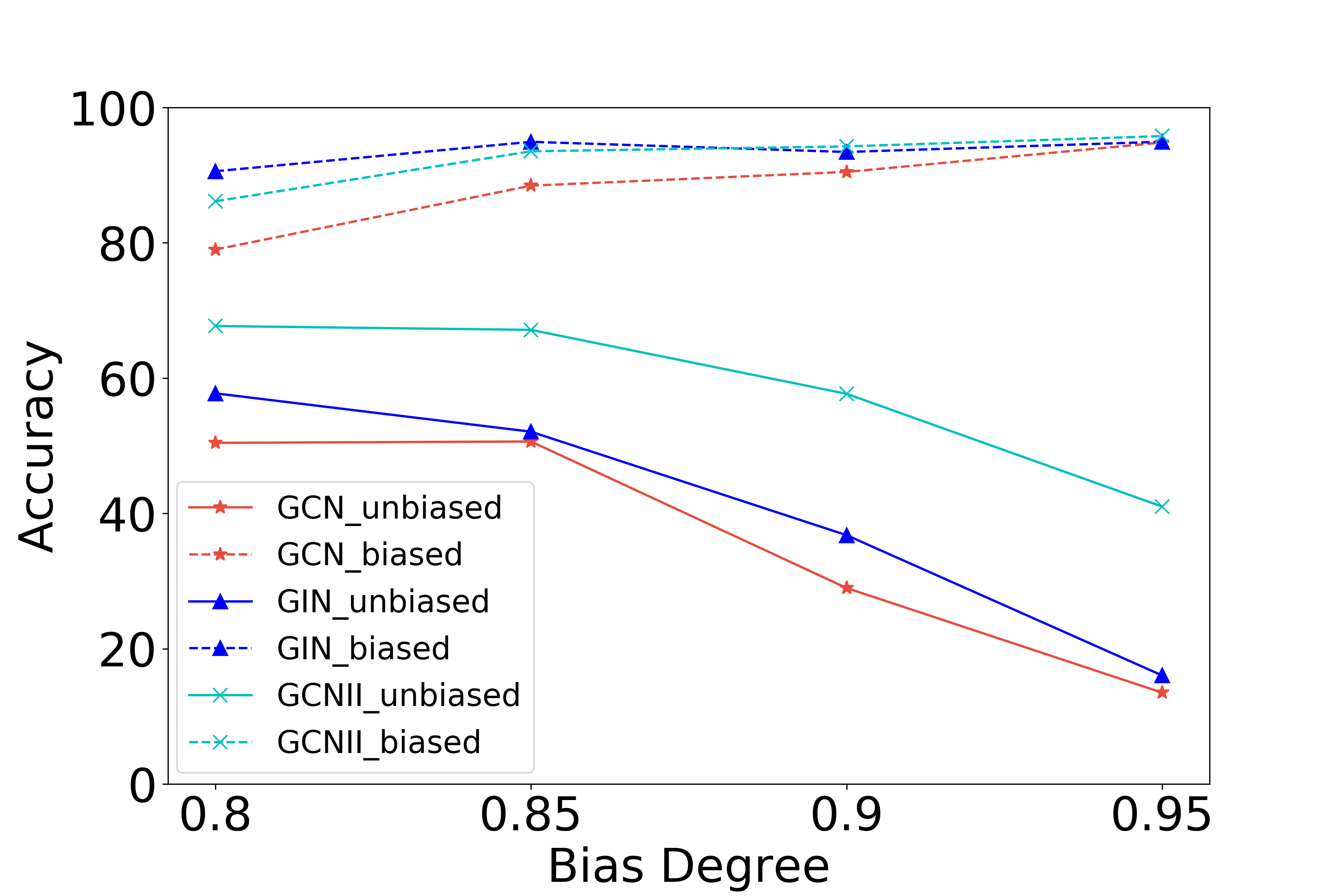}
\label{fig:motivating results}
}
\caption{Example graphs of CMNIST-75sp and the performance of GNNs on this dataset.}
\label{fig:motivating}
\end{figure}

Ideally, when the graphs are unbiased, $i.e.$, only the causal substructures are related with the graph labels, the GNNs are able to utilize such substructure to predict the labels. However, due to the uncontrollable data collection process, the graphs are inevitably biased, $i.e.$, existing meaningless substructures spuriously correlates with labels. Taking a colored MNIST superpixel graph dataset in Sec.~\ref{Sec:motivating example} as an example (illustrated in Fig.~\ref{fig:illustration}), each category of digit subgraphs mainly correspond to one kind of color background subgraphs, $e.g.$, digit 0 subgraph is related with red background subgraph. Therefore, the color background subgraph will be treated as bias information, which highly correlates with labels but does not determines them in the training set. Under this situation, \textit{will GNNs still stably utilize the causal substructure to make decision?}

To investigate the impact of bias on GNNs, we conduct an experimental investigation to demonstrate the impact of bias (especially in the severe bias scenarios) on the generalization capability of GNNs (Sec.~\ref{Sec:motivating example}). We find that GNNs actually utilize both bias and causal substructures to make prediction. However, with severer bias correlation, even bias substructure still could not exactly determine labels like causal substructure, GNNs majorly utilize bias substructure as shortcuts to make prediction, causing a large generalization performance degradation. Why this happens? We analyze the data-generating process and model prediction mechanism behind the graph classification using a causal graph (Sec.~\ref{Sec:causal view}). The casual graph illustrates that the observed graphs are generated by the causal and bias latent variables and existing GNNs could not distinguish the causal substructure from entangled graphs. \textit{How can we disentangle the causal and bias substructures from observed graphs, so that GNNs can only utilize the causal substructures to make stable prediction when severe bias appears?}

\par To address the question, two challenges need to be faced. 1) How to identify the causal substructure and bias substructure in the severe biased graphs? In the severe bias scenarios, bias substructure will be ``easier to learn'' for GNNs and finally dominate the prediction. Using the normal cross-entropy loss, like DIR~\cite{wu2022discovering}, could not fully capture such aggressive property of bias. 2) How to extract the causal substructure from an entangled graph? The statistically causal substructure is usually determined by the global property of the entire graph population, rather than a single graph. When extracting causal substructure from a graph, we need to establish the relations among all the graphs.

\par In this paper, we propose a novel debiasing framework for GNNs via learning \textbf{Dis}entangled \textbf{C}ausal substructure, called \textbf{DisC}. Given an input biased graph, we propose to explicitly filter edges into causal and bias subgraphs by a parameterized edge mask generator, whose parameters are shared across entire graph population.  As a result, the edge masker is naturally capable to specify the importance for each edge and extract causal and bias subgraphs from a global view of the entire observations. Then, a ``casual''-aware (weighted cross-entropy) loss and a ``bias''-aware (generalized cross-entropy) loss are respectively utilized to supervise two functional GNN modules. Based on the supervision, the edge mask generator could generate corresponding subgraphs and the GNNs could encode corresponding subgraphs into their disentangled embeddings. With the disentangled embeddings, we randomly permute the latent vectors extracted from different graphs to generate more unbiased counterfactual samples in embedding space. The new generated samples still contain both causal and bias information, while their correlation has been decorrelated. In this time, there is only correlation between causal variables with labels, so that the model could concentrate on the true correlation between the causal subgraphs and labels. Our major contributions are as follows:

\begin{itemize} [leftmargin = 15 pt]
\item To our knowledge, we first study the generalization problem of GNNs in a more challenging yet practical scenario, i.e., the graphs are with severe bias. We systematically analyze the bias impact on GNNs from both experimental study and causal analysis. We find that the bias substructure, compared with causal substructure, is much easier to dominate the training of GNNs.

\item To debias GNNs, we develop a novel GNN framework for disentangling causal substructure, which is flexible to build upon various GNNs for improving generalization ability while enjoying inherent interpretability, robustness and transferability.

\item We construct three new datasets with various properties and controllable bias degrees, which can better benchmark the new problem. Our model outperforms the corresponding base models with a large margin (from 4.47\% to 169.17\% average improvements). Various investigation studies demonstrate that our model could discover and leverage causal substructure for prediction.
\end{itemize}

\section{Related Works}
\textbf{Generalization of GNNs in wild environments.} Most existing GNN methods are proposed under the IID hypothesis, $i.e.,$ training and
testing set are independently sampled from the identical distribution~\cite{scarselli2008graph,kipf2016semi,velivckovic2017graph,hamilton2017inductive,liao2020pac}. However, in reality, thus ideal assumption is hard to be satisfied. Recently, several methods have been proposed to improve the generalization ability of GNNs in wild OOD environments. Several works~\cite{ma2021subgroup,fan2022debiased,wu2022handling} study the OOD problem of node classification.
For OOD graph classification task, StableGNN~\cite{fan2021generalizing} propose to learn the stable causal relationship in graphs. OOD-GNN~\cite{li2021ood} propose to constrain each dimension of learned embedding independent. DIR~\cite{wu2022discovering} discovers the invariant rationales for generalizing GNNs. Although they have achieved better OOD performance, they are not designed  for the datasets with severe bias, which is more challenging for guaranteeing the generalization ability of GNNs. 

\textbf{Disentangled graph neural networks.} Recently,  there are a couple of methods that study the disentangled GNNs. DisenGCN~\cite{ma2019disentangled} utilizes neighbourhood routing mechanism to divide the neighbours of the node into several mutually exclusive parts. IPGDN~\cite{liu2020independence} promotes DisenGCN by constraining the different parts of the embedding feature independent. DisenGCN and IPGDN are node-level disentanglement, thus FactorGCN~\cite{yang2020factorizable} considers the whole graph information and disentangles the target graph into several factorized graphs. Despite results of the previous works, they do not consider disentangling the causal and bias information for graphs.

\textbf{General debiasing methods.} Recently, debiasing problem has drawn much attention in machine learning community~\cite{kim2019learning,li2019repair,sagawa2019distributionally,bahng2020learning,cadene2019rubi,geirhos2018imagenet}. One category of these methods is pre-defining a certain bias type explicitly to mitigate~\cite{kim2019learning,li2019repair,sagawa2019distributionally,bahng2020learning, wang2019learning}. For example, Wang et al.~\cite{wang2019learning} and Bahng et al.~\cite{bahng2020learning}  design a texture- and color-guided  model to adversarially train a debiased neural network against the biased one. Instead of defining certain types of bias, recent approaches~\cite{nam2020learning, darlow2020latent, lee2021learning} rely on the straightforward assumption that models are prone to exploit the bias as shortcuts to make prediction~\cite{geirhos2020shortcut}. In the line with the recent studies, our study belongs to the second category. However, most of existing methods are designed for image datasets and could not effectively extract causal substructure from graph data. Distinctly, we first study the severe bias problem on graph data, and our method could effectively extract causal substructure from graph data.

\section{Preliminary Study and Analysis}
In this section, we first illustrate the existing GNNs tend to exploit the bias substructure as shortcuts for prediction through a motivating experiment. Then we analyze the prediction process of GNNs in causal view. Based on this causal view, it motivates our solution to relieve the impact of bias.
\subsection{Motivating Example}
\label{Sec:motivating example}
To measure the generalization ability of GNNs with the effect of bias, we construct a graph classification dataset with controllable bias degrees, called CMNIST-75sp. We first construct a biased MNIST image dataset like~\cite{bahng2020learning},  where each category of digit highly correlates with a pre-defined color in their background. For example, in the training set, 90\% of 0 digits are with red background ($i.e.,$ biased samples), and remaining 10\% images are with random background color ($i.e.,$ unbiased samples), whose the bias degree is 0.9 in this situation. We consider four bias degrees $\{0.8, 0.85, 0.9, 0.95\}$. For the testing set, we construct both biased testing set and unbiased testing set. The biased testing set has the same bias degree with training set, aiming to measure the extent of models relying on bias.  The unbiased testing set,  where the digit labels uncorrelate with the background colors, aims to test whether the model could utilize the inherent digit signals for prediction. Note that training set and testing set have the same pre-defined color set. Then, we convert the biased MNIST images into superpixel graphs with at most 75 nodes each graph using~\cite{knyazev2019understanding}, where the edges are constructed by the KNN method based on the 2D coordinates of superpixels and node features are the concatenation of coordinates and average color of superpixels. Each graph is labeled by its digit class, so that its digital subgraph is deterministic for label and background subgraph is spuriously correlated with labels but not deterministic. The examples of graphs are illustrated in Fig.~\ref{fig:illustration}.

\par We perform three popular GNN methods: GCN~\cite{kipf2016semi}, GIN~\cite{xu2018how}, and GCNII~\cite{chen2020simple} on CMNIST-75sp and the results are shown in Fig.~\ref{fig:motivating results}. The same color of dashed line and solid line represent the results of the corresponding methods on the biased testing set and the unbiased testing set respectively. Overall, the GNNs achieve much better performance on biased testing set than unbiased testing set. The phenomenon indicates that although GNNs could still learn some causal signals for prediction, the unexpected bias information is also being utilized for prediction. More specifically, with bias degree becoming larger, the performance of GNNs on biased testing set is increased and the value of accuracy is nearly in line with the bias degree, while the performance on unbiased testing drops dramatically. Hence, although causal substructure could determine labels perfectly, in severe bias scenarios, the GNNs lean to utilize the easier to learn bias information to make prediction rather than the inherent causal signals, and bias substructure will finally dominate the prediction.

\begin{figure}[!htbp]
\centering
\subfigure[SCM of the union of the data generation and the existing GNNs' prediction process.]{
\includegraphics[ width=4cm]{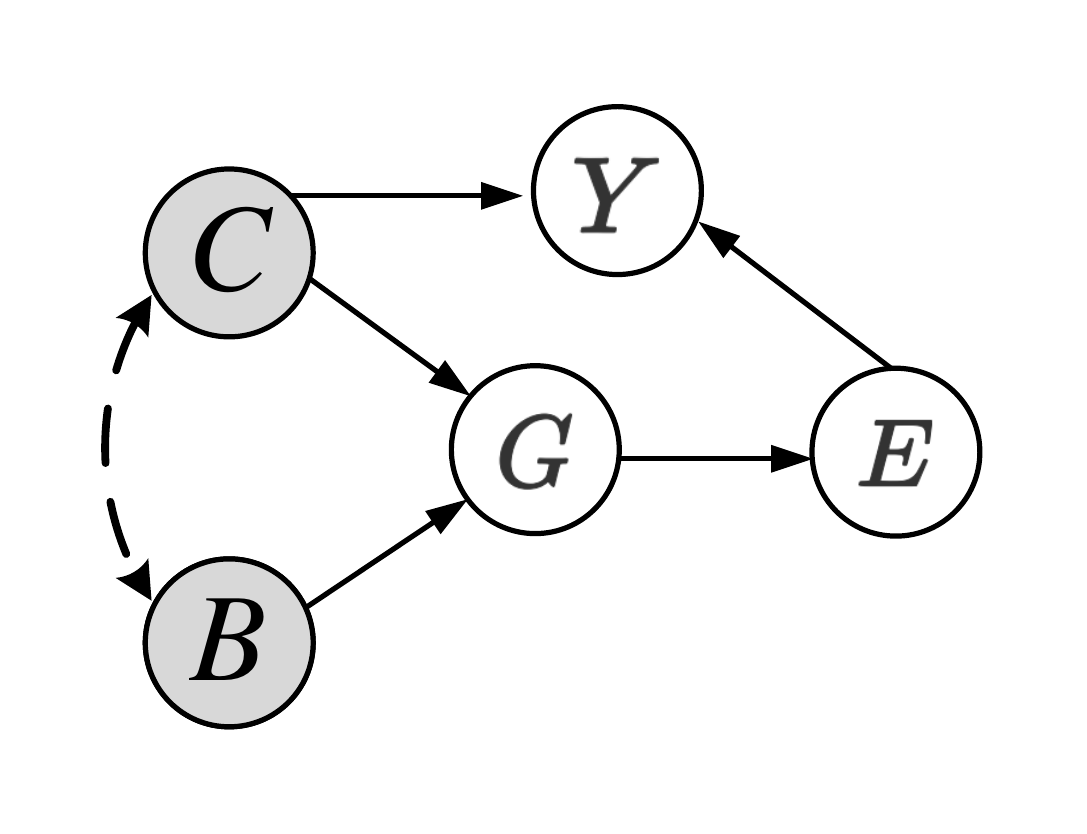}
\label{fig::generation}
}
\subfigure[SCM of our debiasing GNN method.]{
\includegraphics[ width=4cm]{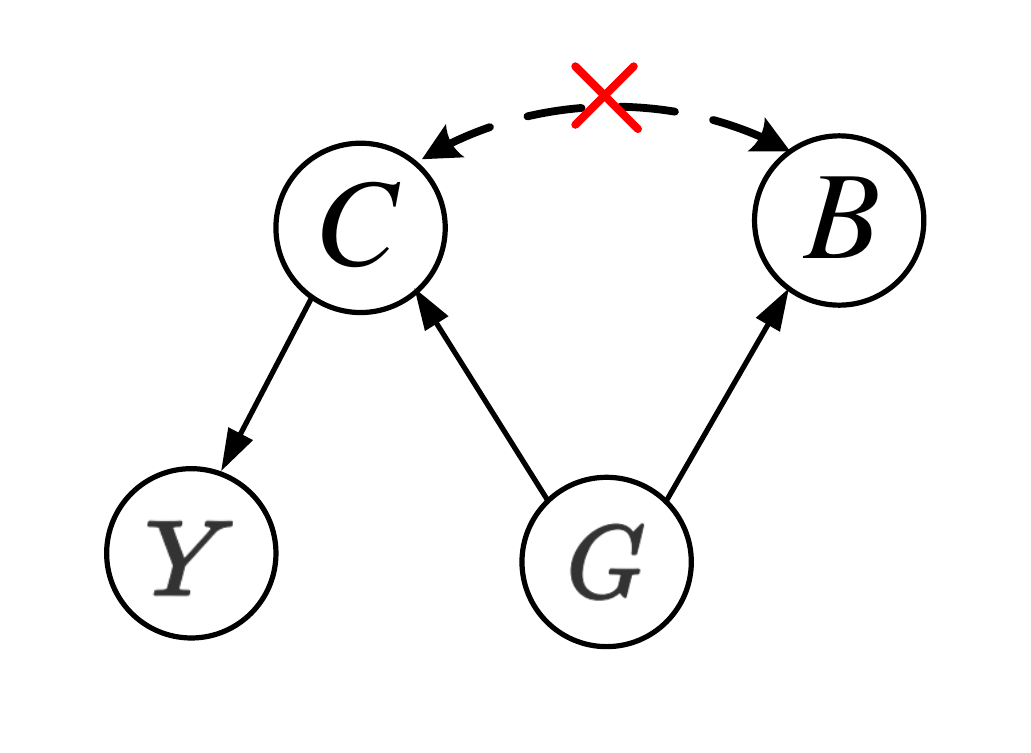}
\label{fig::debiasing}
}
\caption{SCMs. Grey and white variables represent unobserved and observed variables, respectively.}
\label{fig::causal view}
\end{figure}

\subsection{Problem Analysis}
\label{Sec:causal view}
Debiasing GNNs for unbiased prediction requires understanding the natural mechanisms of graph classification task. We present a causal view of the union of the data-generating process and the model prediction process behind the task. Here we formalize the causal view as a Structure Causal Model (SCM) or causal graph~\cite{glymour2016causal,pearl2009causality} by inspecting on the causalities among five variables: unobserved causal variable $C$, unobserved  bias variable $B$,  observed graph $G$, graph embedding $E$, and ground truth label / prediction $Y$\footnote{We use variable Y for both the ground-truth labels and prediction, as they are optimized to be the same.}. Fig.~\ref{fig::generation} illustrates the SCM, where each link denotes a causal relationship.
\begin{itemize} [leftmargin = 15 pt]
\item $C\rightarrow G \leftarrow B$. The observed graph data is generated by two unobserved latent variables: the causal variable $C$ and the bias variable $B$, such as digit subgraphs and background subgraphs in the CMNIST-75sp dataset. And all bellow relations are illustrated by CMNIST-75sp.

\item $C\rightarrow Y$. This link means that the causal variable $C$ is the only endogenous parent to determine the generation of ground-truth label $Y$. For example, $C$ is the oracle digit subgraph, which exactly explains why the label is labeled as $Y$.

\item $C\dashleftarrow\dashrightarrow B$. This link indicates the spurious correlation between $C$ and $B$. Such probabilistic dependencies is usually caused by the direct cause or unobserved confounder~\cite{reichenbach1991direction}. Here we do not distinguish these scenarios and only observe the spurious correlation between $B$ and $C$, such as the spurious correlation between the color background subgraphs and digit subgraphs.

\item $G \rightarrow E \rightarrow Y$. Existing GNNs usually learn the graph embedding $E$ based on the observed graph $G$ and make the prediction $Y$ based on the learned embedding $E$.
\end{itemize} 
\par According to the SCM, GNNs will utilize both information to make prediction. As bias substructure ($e.g.,$ background subgraph) usually has simpler structure than meaningful causal substructure ($e.g.,$ digit subgraph), if GNN utilizes such simple substructure, it could achieve low loss very fast. Hence, GNN inclines to utilizes bias information when most graphs are biased. Based on the SCM  in Fig.~\ref{fig::generation}, according to $d$-connection theory~\cite{pearl2009causality} (see App.~\ref{Sec::preli}): two variables are dependent if they are connected by at least one unblocked path, we could find two paths that would induce the spurious correlation between the bias variable $B$ and label $Y$: \textbf{(1) $\mathbf{B\rightarrow G \rightarrow E \rightarrow Y}$} and  \textbf{(2)  $\mathbf{B\leftrightarrow C \rightarrow Y}$}. To make the prediction $Y$ being uncorrelated with the bias $B$, we need to intercept the two unblocked paths. For this purpose, we propose to debias GNNs in causal view, as in Fig.~\ref{fig::debiasing}.
\begin{itemize}[leftmargin = 15 pt]
\item $C\leftarrow G \rightarrow B$ and $C\rightarrow Y$. To intercept the path (1), we should disentangle the latent variables $C$ and $B$ from the observed graph $G$ and make prediction only based on the causal variable $C$.

\item $C\dashleftarrow\Ccancel{ -- }\dashrightarrow B$. To intercept the path (2), as we cannot change the link between $C$ and $Y$, one possible solution is to make $C$ and $B$ uncorrelated.
\end{itemize}

\section{Methodology}
\begin{figure*}[!htbp]
  \centering
  \includegraphics[width=13cm]{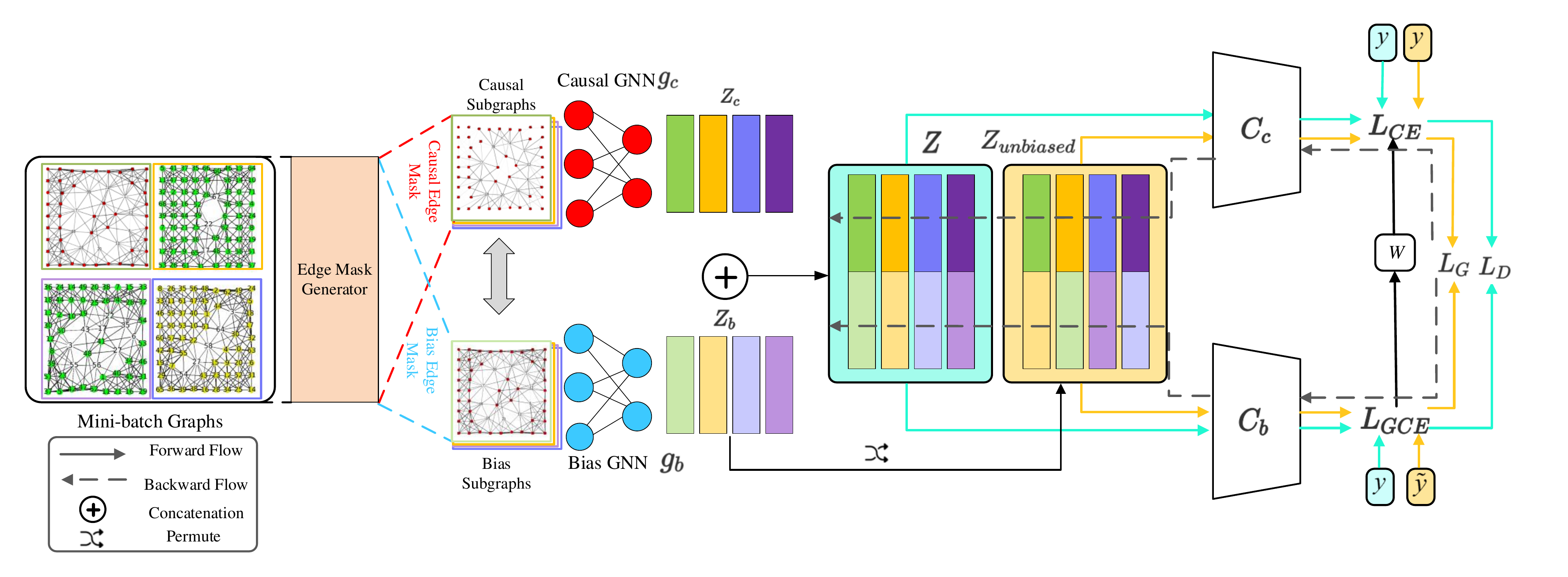}
  \caption{The overall framework of DisC. }
  \label{fig::Model}
\end{figure*}

Motivated by the above causal analysis, in this section, we present our proposed debiasing GNN framework \textbf{DisC},  to remove the spurious correlation. The overall framework is shown in Fig.~\ref{fig::Model}. First, an edge mask generator is learnt to mask the edges of original input graphs into causal subgraphs and bias subgraphs. Second, two separate GNN modules with their corresponding masked subgraphs are trained to encode corresponding causal substructure and bias substructure into disentangled representations, respectively. Last, after the disentangled representations are well-trained, we 
permute the bias representations among the training graphs to generate counterfactual unbiased samples, so that the correlation between causal representations and bias representations is removed.
\subsection{Causal and Bias Substructure Generator}
Given a mini-batch of biased graphs $\mathcal{G}=\{G_1, \cdots, G_n\}$, our idea is that: we take a collection of graph instances and design a generative probabilistic model to learn to mask the edges into causal subgraph or bias subgraph. Particularly, given a graph $G=\{\mathbf{A}, \mathbf{X}\}$, where  $\mathbf{A}$ is the adjacency matrix and $\mathbf{X}$ is the node feature matrix, we utilize a multi-layer perceptron (MLP) upon the concatenation of node features $\mathbf{x}_i$ of node $i$ and $\mathbf{x}_j$ of node $j$ to measure the importance of edge $(i, j)$ for causal subgraph: 
\begin{equation}
\quad \alpha_{ij} = \text{MLP}([\mathbf{x}_i, \mathbf{x}_j]).
\end{equation}
Then a sigmoid function $\sigma(\cdot)$ is employed to project $\alpha_{ij}$ into the range of (0,1), which indicates the probability of edge $(i, j)$ being the edge in the causal subgraph as follows:
\begin{equation}
    c_{ij}=\sigma(\alpha_{ij}).
\end{equation}
Naturally, we could get the probability of edge $(i, j)$ being the edge in the bias subgraph by: $b_{ij}=1-c_{ij}.$ Now we could construct the causal edge mask $\mathbf{M}_c=[c_{ij}]$ and bias edge mask $\mathbf{M}_b=[b_{ij}]$. Finally, we decompose the original graph $G$ into causal subgraph $G_c=\{\mathbf{M}_c\odot \mathbf{A}, \mathbf{X}\}$ and bias subgraph $G_b=\{\mathbf{M}_b\odot \mathbf{A}, \mathbf{X}\}$. Intuitively, the edge mask could highlight different part of structure information of original graphs, thus GNNs built on the different subgraphs could encode different parts of graph information. Moreover, the mask generator has two advantages. (1) \textbf{Global view:} In individual graph level, the mask generator ($i.e.$, MLP), whose parameters are shared by all the edges in a graph, take a global view of all the edges in a graph, which enables us to identify community in graph. It is well known that the effect of an edge cannot be judged independently, because edges usually collaborate with each other, forming a community, to make prediction. Thus, it is critical to evaluate an edge in a global view. In whole graph population level, the mask generator takes a global view of all the graphs in the training set, which enables us to identify causal/bias subgraph. Particularly, as the causal/bias is the statistical information in the population level, it is necessary to view all the graphs to identify the causal/bias substructure. Considering both such coalition effects and population-level statistical information, the generator is able to measure the importance of edges more accurately. (2) \textbf{Generalization}: The mask generator can generalize the mechanism of mask generation to new graphs without retraining, so it is capable and efficient to prune unseen graphs.

\subsection{Learning Disentangled Graph Representations}
Given $G_c$ and $G_b$, how to ensure they are causal subgraph and bias subgraph, respectively? Inspired by~\cite{lee2021learning}, our approach simultaneously trains a pair of GNNs $(g_b, g_c)$ with linear classifiers $(C_b, C_c)$ as follows: (1) Motivated by the observation in Sec.~\ref{Sec:motivating example} that bias substructure is easier to learn, we utilize a bias-aware loss to train a bias GNN $g_b$ and a bias classifier $C_b$ and (2) in contrast, we train a causal GNN $g_c$ and a causal classifier $C_c$ on the training graphs that the bias GNN struggles to learn. Next, we would present each component in detail.

As shown in Fig.~\ref{fig::Model}, GNN $g_c$ and $g_b$ embed the corresponding subgraphs into causal embedding $z_c=g_c(G_c; \gamma_c)$ and bias embedding  $z_b=g_b(G_b; \gamma_b)$, respectively, where $\gamma$ is the parameters of GNNs. Subsequently, concatenated vector $z=[z_c; z_b]$ is fed into linear classifiers $C_c$ and $C_b$ to predict the target label $y$. To train $g_b$ and $C_b$ as bias extractor, we utilize the generalized cross entropy (GCE)~\cite{zhang2018generalized} loss to amplify the bias of the bias GNN and classifier:
\begin{equation}
    GCE(C_b(z;\alpha_b),y) = \frac{1-C_b^y(z;\alpha_b)^q}{q},
\end{equation}
where $C_b(z;\alpha_b)$ and $C_b^y(z;\alpha_b)$ are softmax output of the bias classifier and its probability belonging to the target category $y$, respectively, and $\alpha$ is the parameters of classifier. Here $q\in(0,1]$ is a hyperparameter that controls the degree of amplifying bias. Given $\theta_b = [\gamma_b, \alpha_b]$, the gradient of the GCE loss up-weights the gradient of the standard cross entropy (CE) loss for the samples with a high confidence $C_b^y$ of predicting the correct target category as follows: 
\begin{equation}
    \frac{\partial GCE(C_b(z;\alpha_b),y)}{\partial\theta_b}={(C_b^{y})^q}\frac{\partial CE(C_b(z;\alpha_b),y)}{\partial\theta_b}.
    \label{Eq:gradient}
\end{equation}
Therefore, compared with CE loss, GCE loss will amplify the gradients of $\theta_b$ on samples by the confidence score $(C_b^{y})^q$. Based on our observation that the bias information is usually easier to be learned, so the biased graphs will have higher $(C_b^{y})^q$ than unbiased graphs. Therefore, the model $g_b$ and $C_b$ trained by GCE loss will focus on bias information and finally get the bias subgraph. Note that, to ensure that $C_b$ predicts target labels mainly based on this $z_b$, the loss from $C_b$ is not backpropagated to $g_c$, $i.e.,$ only update $\theta_b$ in Eq. (\ref{Eq:gradient}), and vice versa.

\par Meanwhile, we also train a causal GNN simultaneously with the weighted CE loss. The graphs with high CE loss from $C_b$ can be regarded as the unbiased samples compared with the samples with low CE loss. In this regard, we could obtain the unbias score of each graph as
\begin{equation}
    W(z) = \frac{CE(C_b(z), y)}{CE(C_c(z), y)+CE(C_b(z), y)}.
\end{equation}
Large value of $W$ implies the graph is an unbiased sample, hence we could use these weights to reweight the loss of these graphs to train $g_c$ and $C_c$, enforcing them to learn the unbiased information. Thus, the objective function for learning disentangled representation is:
\begin{equation}
    L_D = W(z) CE(C_c(z), y) + GCE(C_b(z), y).
    \label{Eq:LD}
\end{equation}

\subsection{Counterfactual Unbiased Sample Generation}
\par Until now, we have achieved the first goal analyzed in Sec.~\ref{Sec:causal view} that is the disentanglement of causal and bias substructures. Next, we will show how to achieve the second goal that makes the causal variable $z_c$ and bias variable $z_b$ uncorrelated. Although we have disentangled causal and bias information, they are disentangled from the biased observed graphs. Hence, there will exist statistical correlation between causal and bias variables inheriting from the biased observed graphs. To further decorrelate $z_c$ and $z_b$, according to the causal relation of data-generating process: $C\rightarrow G\leftarrow B$, we propose to generate the counterfactual unbiased samples in embedding space by swapping $z_b$. More specifically, we randomly permute bias vectors in each mini-batch and obtain $z_{unbiased}=[z_c;\hat{z_b}]$, where $\hat{z_b}$ represents the randomly permuted bias vectors of $z_b$. As $z_c$ and $\hat{z_b}$ in $z_{unbiased}$ are randomly combined from different graphs, they will have much less correlation than $z=[z_c; z_b]$ where both are from the same graph. To make $g_b$ and $C_b$ still focus on the bias information, we also swap label $y$ as $\hat{y}$ along with $\hat{z_b}$, so that the spurious correlation between $\hat{z_b}$ and $\hat{y}$ still exists. With the generated unbiased samples, we utilize the following loss function to train two GNN modules:
\begin{equation}
    L_{G} = W(z) CE(C_c(z_{unbiased}), y) + GCE(C_b(z_{unbiased}), \hat{y}),
\end{equation}
Together with the disentanglement loss, total loss function is defined as:
\begin{equation}
    L=L_D + \lambda_{G} L_{G},
    \label{Eq:L}
\end{equation}
where $\lambda_{G}$ is a hyperparameter for weighting the importance of generation component. Moreover, training with more diverse samples would also benefit with better generalization on unseen testing scenarios.
 Our approach is summarized in App.~\ref{sec::alg}.
Note that, as we need well-disentangled representations to generate the high-quality unbiased samples, in the early stage of training, we only train the model with $L_D$. After certain epochs, we train the model with $L$.
% \usepackage{multirow}
% \usepackage{multirow}

\section{Experiment}
\label{Sec:Exp}
%This section validates the effectiveness of proposed DisC in debiasing with both quantitative and qualitative evaluation. 

%We compare our method with popular GNN base models and the ablation study could also validate the importance of each proposed component. For the qualitative evaluation, we verify our approach could leverage causal substructure for prediction by visualizing the edge mask. And the visualization of causal embedding and bias embedding also validates that our framework could disentangle them from the biased graphs. 

\textbf{Datasets.} We construct three datasets with various properties and bias ratios to benchmark this new problem, where the datasets have clear causal subgraphs making the results explainable. Following CMNIST-75sp introduced in Sec.~\ref{Sec:motivating example}, we use the similar way to construct CFashion-75sp and  CKuzushiji-75sp datasets based on the Fashion-MNIST~\cite{xiao2017fashion} and Kuzushiji-MNIST~\cite{clanuwat2018deep} datasets.  As the causal subgraphs of these two datasets are more complicated (fashion product and hiragana characters), they are more challenging.  Due to the page limits, here we set bias degrees as $\{0.8, 0.9, 0.95\}$. We report the main results on unbiased test sets.  Details are in App.~\ref{appendix datasets}.

\textbf{Baselines and experimental setup.} As DisC is a general framework which could be built on various base GNN models, we select three popular GNNs:  GCN~\cite{kipf2016semi}, GIN~\cite{xu2018how}, and GCNII~\cite{chen2020simple}. The corresponding models are termed as $\text{DisC}_{GCN}$, $\text{DisC}_{GIN}$ and $\text{DisC}_{GCNII}$, respectively. Hence, base models are the most straight baselines. Another kind fo baselines are the causal-inspired GNN method DIR~\cite{wu2022discovering} and StableGNN~\cite{fan2021generalizing}. We also compare against a general debiasing method LDD~\cite{lee2021learning} by replacing its encoder with GNNs. Graph Pooling method DiffPool~\cite{ying2018hierarchical} and graph disentangling method FactorGCN~\cite{yang2020factorizable} are also compared. To keep fair comparison, our model uses the same GNN architecture and hyperparameters with the corresponding base model. All the experiments are run 4 times with different random seeds and we report the accuracy and the standard error. More details are in App.~\ref{appendex setup}.

\subsection{Quantitative Evaluation}
\label{Sec::quan}
\textbf{Main results.} The overall results are summarized in Table~\ref{tab:main}, and we have following observations:
\begin{itemize}  [leftmargin = 15 pt]
    \item [(1)]DisC has much better generalization ability than base models. DisC outperforms the corresponding base model consistently with a large margin. With heavier biases, our model achieves larger improvements over base models. Specifically, for CMNIST-75sp, CFashion-75sp and CKuzushiji-75sp with smaller bias degree ($i.e.,$ 0.8), our model achieves 40.02\%, 4.47\% and 29.82\% average improvements over corresponding base models, respectively. Surprisingly, with severer biases (0.9 and 0.95), DisC achieves 169.17\%, 14.67\% and 49.35\% average improvements over base models on three datasets, respectively. It indicates that the proposed method is a general framework helping existing GNNs against the negative impact of bias.
    
    %And we notice that DIR could not achieve satisfying results. The reason is that this method is not designed for the severe bias scenarios. Such improvements strongly validate that DisC is an effective framework to remove the unexpected bias in the training set.
    \item [(2)]DisC significantly outperforms existing debiasing methods. We notice that DIR could not achieve satisfying results. The reason is that DIR utilizes CE loss to extract bias information, which could not fully capture the property of bias in severe bias scenarios. And DIR sets one fixed threshold to spilt subgraphs, which is suboptimal. StableGNN outperforms their base model DiffPool and achieve competitive results, indicating the effectiveness of their proposed causal variable distinguishing regularizer. However, their framework adjusts data distribution based on the original dataset,  it is hard to generate unbiased distribution when the unbiased samples are scarce. DisC could generate more unbiased samples based on the disentangled representations. Moreover, LDD is a general debiasing method which is not designed for graph data. DisC outperforms corresponding LDD variants with average 23.15\%, indicating that the seamless joint of global-population-aware edge masker with debiasing disentangle framework is very effective.
    %\item [(2)]\textbf{Learning edge mask is extremely important for removing the bias information.} As we can see from the results, our model with mask variants (DisC/w. M w.o. G and DisC/w. M w. G) averagely surpasses the corresponding without mask variants (DisC/w.o. M w.o. G and DisC/w.o. M w. G) by 21.14\%. This phenomenon validates that explicitly learning substructure is crucial for GNNs to extract disentangled causal information for graph classification task.
    
    %\item [(3)]\textbf{Generating unbiased samples further boosts model performance.} From the results, we could observe that the variants with generation (DisC/w. M w. G) significantly outperform the corresponding variants without generation (DisC/w. M w.o. G) with average 6.46\%. It indicates that generating counterfactual unbiased samples in embedding space will further remove the correlation between bias variables and prediction, conforming with causal analysis in Section~\ref{Sec:causal view}.
\end{itemize}

\begin{table}[!htbp]
\caption{Graph classification accuracy evaluated on unbiased testing sets, which have same color (bias) set with training set. The best performance within each base model variant is in bold.}
\label{tab:main}
\resizebox{1\textwidth}{!}{
 \centering
\begin{tabular}{l|lll|lll|lll}

\hline
\multicolumn{1}{c|}{Dataset}                                         & \multicolumn{3}{c|}{CMNIST-75sp}                                                                                                                        & \multicolumn{3}{c}{CFashion-75sp}     & \multicolumn{3}{|c}{CKuzushiji-75sp}                                                                                        \\ 
\multicolumn{1}{c|}{Bias}                                            & \multicolumn{1}{c}{0.8}               & \multicolumn{1}{c}{0.9}                & \multicolumn{1}{c|}{0.95} & \multicolumn{1}{c}{0.8}                             & \multicolumn{1}{c}{0.9}                 & \multicolumn{1}{c|}{0.95} & \multicolumn{1}{c}{0.8}                             & \multicolumn{1}{c}{0.9}                 & \multicolumn{1}{c}{0.95} \\ \hline
FactorGCN~\cite{yang2020factorizable}                                            & $72.30_{\pm 1.18}$                     & $62.35_{\pm 5.07}$                 &    $42.50_{\pm 4.91}$                        & $61.23_{\pm 1.11}$  &$53.50_{\pm 1.29}$ &$45.78_{\pm 2.40}$    &  $42.87_{\pm 1.19}$   &      $32.35_{\pm 2.79}$ & $23.87_{\pm 0.12}$                \\
DiffPool~\cite{ying2018hierarchical}                                             & $73.79_{\pm 0.02}$                      & $66.45_{\pm 0.78}$                  &     $47.12_{\pm 1.04}$                       & $62.82_{\pm 0.53}$   &$57.50_{\pm 0.39}$ &$50.86_{\pm 0.20}$     &     $45.46_{\pm 0.65}$   &      $36.18_{\pm 0.19}$  & $27.45_{\pm 0.26}$                 \\ \hline
DIR~\cite{wu2022discovering}                                             & $9.98_{\pm 0.33}$                     & $9.96_{\pm 0.23}$                   &     $10.03_{\pm 0.27}$                        & $13.02_{\pm 1.92}$  &$12.80_{\pm 1.67}$ &$11.98_{\pm 1.41}$     &     $10.35_{\pm 0.32}$   &      $10.72_{\pm 0.27}$   & $10.59_{\pm 0.46}$                  \\
StableGNN~\cite{fan2021generalizing}                                            & $77.65_{\pm 1.64}$                      & $68.87_{\pm 1.74}$                   &     $51.33_{\pm 0.87}$                        & $64.03_{\pm 0.29}$   &$58.26_{\pm 0.09}$ &$51.46_{\pm 0.39}$     &      $49.41_{\pm 0.09}$   &      $39.30_{\pm 0.12}$   & $28.26_{\pm 0.14}$                  \\
$\text{LDD}_{GCN}$~\cite{lee2021learning}  & $64.95_{\pm 1.22}$   & $56.65_{\pm 2.18}$ & $46.83_{\pm 2.88}$        & $63.85_{\pm 1.17}$  & $64.30_{\pm 0.89}$   & $62.28_{\pm 0.48}$ & $42.38_{\pm 0.33}$ & $38.75_{\pm 0.49}$ & $33.08_{\pm 0.59}$ \\
$\text{LDD}_{GIN}$~\cite{lee2021learning}   &  $64.88_{\pm 1.45}$ & $50.59_{\pm 1.07}$ & $31.23_{\pm 2.48}$        & $64.65_{\pm 0.63}$ & $57.10_{\pm 0.43}$  & $53.38_{\pm 0.47}$ &   $37.83_{\pm 0.54}$ & $28.97_{\pm 0.18}$ & $22.13_{\pm 0.34}$ \\
$\text{LDD}_{GCNII}$~\cite{lee2021learning}   &  $78.03_{\pm 0.66}$ & $69.53_{\pm 0.96}$ & $51.05_{\pm 3.87}$        & $50.63_{\pm 1.79}$ & $54.09_{\pm 2.54}$ & $57.93_{\pm 0.88}$  & $48.70_{\pm 1.98}$  & $41.59_{\pm 1.07}$  & $33.93_{\pm 0.71}$ \\
\hline
   
 GCN~\cite{kipf2016semi}                & $50.43_{\pm 4.13}$ & $28.97_{\pm 4.4}$  & $13.50_{\pm 1.38}$        & $63.60_{\pm 0.53}$  & $57.22_{\pm 0.93}$  & $47.69_{\pm 0.42}$  &  $38.45_{\pm 1.1}$ & $28.35_{\pm 0.79}$ &$20.70_{\pm 0.88}$ \\% & $45.19_{\pm 0.65}$      \\  

 $\text{DisC}_{GCN}$   & $\mathbf{82.60_{\pm 0.93}}$   & $\mathbf{78.14_{\pm 2.14}}$ & $\mathbf{63.47_{\pm 5.65}}$        & $\mathbf{66.85_{\pm 1.11}}$  & $\mathbf{65.33_{\pm 4.70}}$  & $\mathbf{63.93_{\pm 1.50}}$ & $\mathbf{55.53_{\pm 2.29}}$ & $\mathbf{48.13_{\pm 2.59}}$ &  $\mathbf{36.63_{\pm 1.73}}$\\ \hline% &  $59.20_{\pm 1.57}$        \\ 
GIN~\cite{xu2018how}                & $57.75_{\pm 0.78}$ & $36.78_{\pm 5.55}$ & $16.04_{\pm 1.14}$        & $64.25_{\pm 0.46}$ & $58.03_{\pm 0.40}$  & $49.74_{\pm 0.60}$ &  $41.83_{\pm 0.78}$ &   $30.09_{\pm 0.87}$ &   $21.18_{\pm 1.63}$    \\ %&$70.44_{\pm 1.18}$     \\ 
  
  $\text{DisC}_{GIN}$     & $\mathbf{82.10_{\pm 1.50}}$   & $\mathbf{74.90_{\pm 1.81}}$  & $\mathbf{58.58_{\pm 4.24}}$        & $\mathbf{67.10_{\pm 1.07}}$ & $\mathbf{59.90_{\pm 1.31}}$  & $\mathbf{55.80_{\pm 0.36}}$& $\mathbf{55.18_{\pm 1.00}}$  & $\mathbf{41.75_{\pm 0.81}}$& $\mathbf{30.25_{\pm 1.63}}$ \\ \hline% & $70.58_{\pm 0.90}$      \\ \hline
  
GCNII~\cite{chen2020simple}             & $69.70_{\pm 1.73}$   & $57.68_{\pm 1.68}$ & $41.00_{\pm 3.75}$         & $\mathbf{66.68_{\pm 0.59}}$ & $60.58_{\pm 0.28}$  & $53.18_{\pm 0.08}$   & $48.53_{\pm 0.25}$ & $36.23_{\pm 0.20}$  & $25.60_{\pm 0.76}$ \\%  &$68.22_{\pm 0.59}$       \\ 

$\text{DisC}_{GCNII}$     & $\mathbf{79.50_{\pm 2.48}}$ & $\mathbf{76.00_{\pm 1.90}}$ & $\mathbf{60.54_{\pm 5.33}}$        & $66.47_{\pm 1.77}$  & $\mathbf{65.48_{\pm 0.70}}$ & $\mathbf{61.75_{\pm 0.27}}$ & $\mathbf{54.90_{\pm 1.30}}$  & $\mathbf{44.73_{\pm 1.55}}$& $\mathbf{36.95_{\pm 0.70}}$ \\ \hline%  &$\mathbf{71.69_{\pm 1.09}}$    \\ \hline
\end{tabular}}
\end{table}

\begin{figure}[!htbp]
	\centering
	\subfigure[CMNIST-75sp]{
		\includegraphics[ width=4cm]{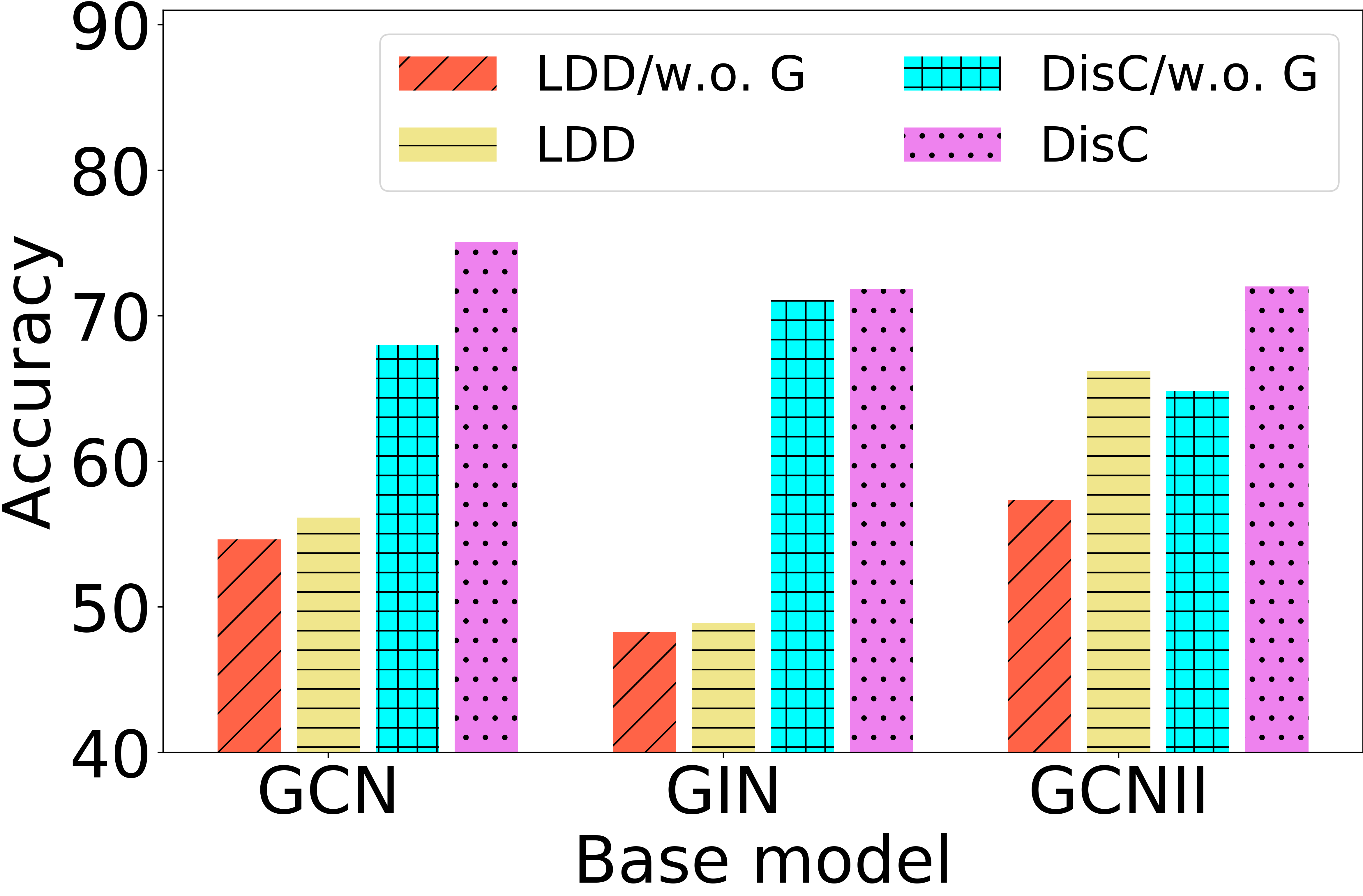}
		\label{fig:ablation_GIN}
	}
	\hspace{10pt}
	\subfigure[CFashion-75sp]{
		\includegraphics[ width=4cm]{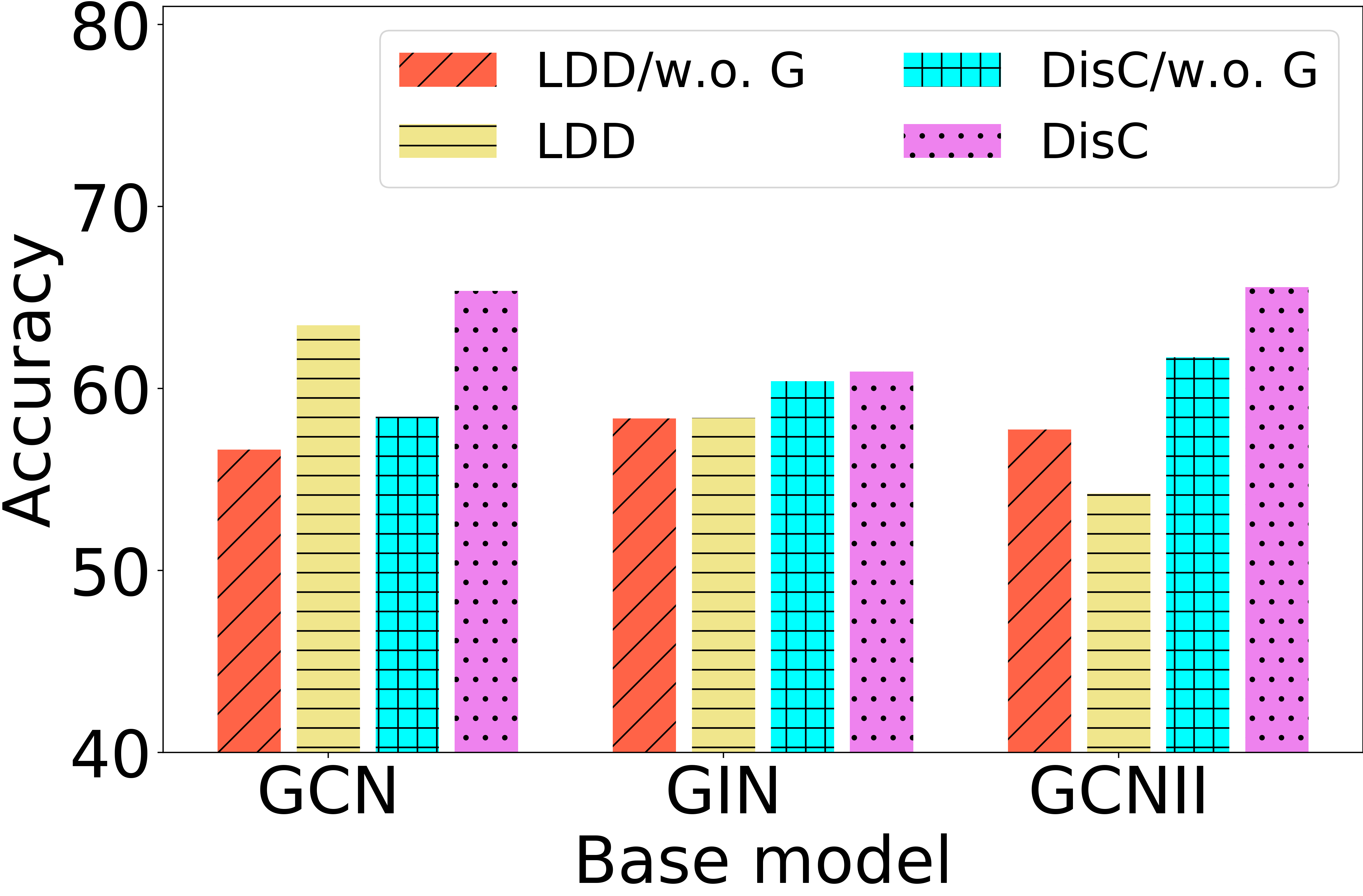}
		\label{fig:ablation_GAT}
	}
	\hspace{10pt}
	\subfigure[CKuzushiji-75sp]{
		\includegraphics[ width=4cm]{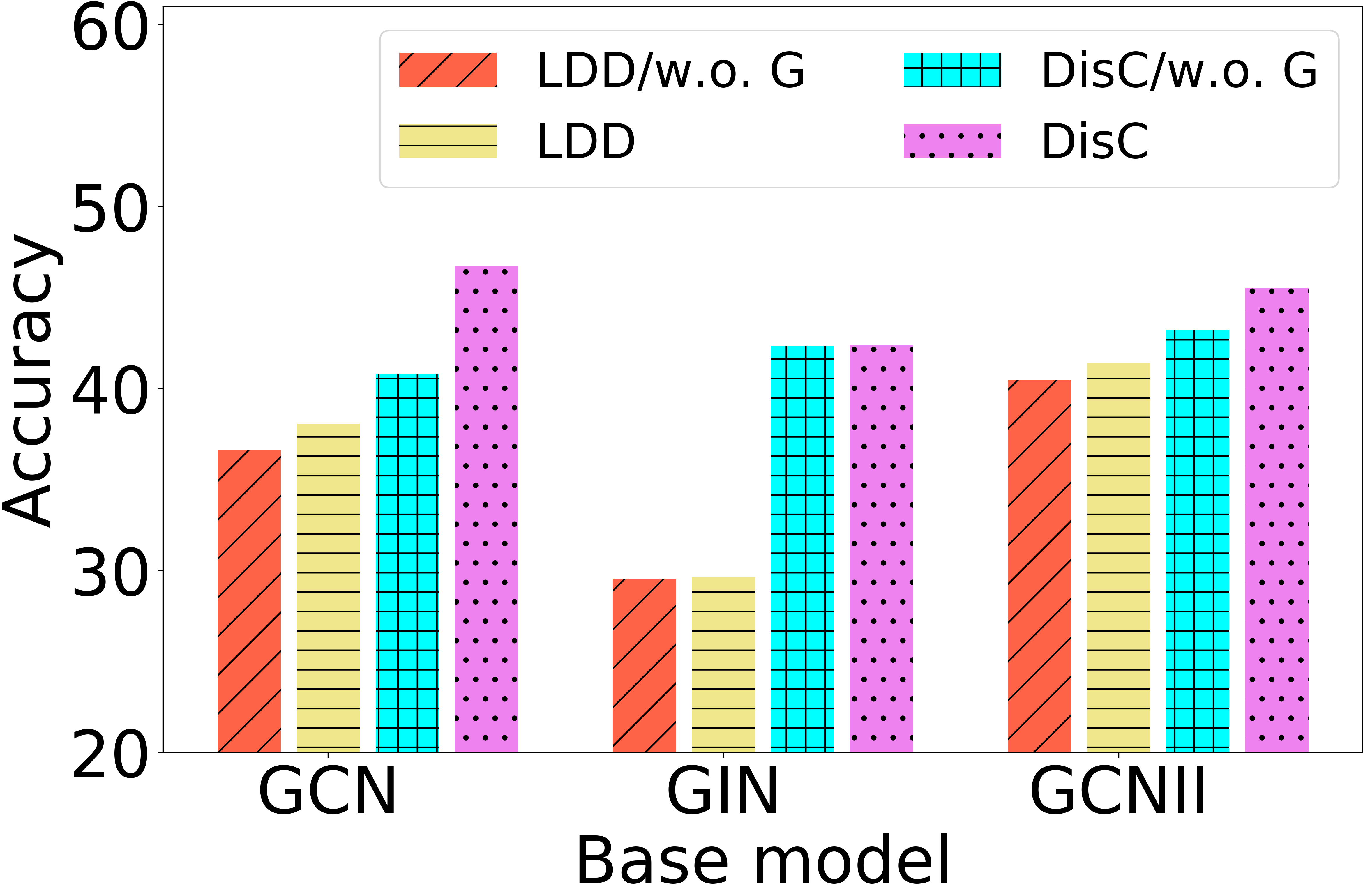}
		\label{fig:ablation_GCNII}
	}
	\caption{Ablation studies of the DisC vs. LDD average over three bias degrees of each dataset.}
	\label{fig:Ablation}
\end{figure}

\textbf{Ablation studies.} To validate the importance of each module in our method, in Fig.~\ref{fig:Ablation}, we conduct ablation studies on our variants (w.o. G means without the sample generation module) and the related variants of LDD. The major difference between DisC/w.o. G with LDD /w.o. G is the edge mask module. In most cases, DisC/w.o. G significantly outperforms LDD /w.o. G, indicating the necessity of learning edge mask for graph data. And DisC which has counterfactual sample generation module could further boost the performances based on the disentangled embeddings of DisC/w.o. G. However, LDD seldomly outperforms LDD /w.o. G or even achieves worse performances. That is, generating high-quality counterfactual samples needs well-disentangled causal and bias embeddings. If embeddings are not well-disentangled, counterfactual samples may act as noisy samples, which would prevent models from achieving further improvement. The edge masker could help the model generate well-disentangled embeddings, which is crucial for overall performance.

\textbf{Robustness on unseen bias.} Table~\ref{tab:unseen} reports the results of DisC compared with its corresponding base models on testing set with unseen bias, $i.e.,$ the pre-defined color (bias) sets of training set and testing set are disjoint. The performances of base models further drop compared with the results on seen bias scenario in Table~\ref{tab:main}. However, our model still achieves very stable performances, fully demonstrating the generalization ability of our model on agnostic bias scenario.

\begin{table}[!htbp]
\setlength\tabcolsep{2pt}
\caption{The results on unseen unbiased testing sets, $i.e.$, the color has not been seen in training set.}
\label{tab:unseen}
\resizebox{1\textwidth}{!}{

 \centering
\begin{tabular}{l|lll|lll|lll}
\hline
\multicolumn{1}{c|}{Dataset}                                         & \multicolumn{3}{c|}{CMNIST-75sp}                                                                                                                        & \multicolumn{3}{c}{CFashion-75sp}     & \multicolumn{3}{|c}{CKuzushiji-75sp}                                                                                        \\ 
\multicolumn{1}{c|}{Bias}                                            & \multicolumn{1}{c}{0.8}               & \multicolumn{1}{c}{0.9}                & \multicolumn{1}{c|}{0.95} & \multicolumn{1}{c}{0.8}                             & \multicolumn{1}{c}{0.9}                 & \multicolumn{1}{c|}{0.95} & \multicolumn{1}{c}{0.8}                             & \multicolumn{1}{c}{0.9}                 & \multicolumn{1}{c}{0.95} \\ \hline
DIR     & $10.38_{\pm 0.28}$  & $10.14_{\pm 0.40}$ & $9.77_{\pm 0.18}$  & $16.77_{\pm 1.71}$ & $16.51_{\pm 3.20}$ & $12.59_{\pm 1.61}$ & $10.48_{\pm 0.34}$ & $10.33_{\pm 0.75}$ & $10.59_{\pm 0.95}$  \\ \hline
GCN     & $36.88_{\pm 5.16}$  & $23.07_{\pm 4.07}$ & $11.88_{\pm 0.33}$  & $59.33_{\pm 0.55}$ & $53.65_{\pm 0.47}$ & $45.60_{\pm 1.06}$ & $36.35_{\pm 0.48}$ & $27.88_{\pm 0.94}$ & $19.95_{\pm 0.67}$  \\ 

DisC$_{GCN}$     & $\mathbf{82.73_{\pm 1.31}}$ & $\mathbf{77.70_{\pm 0.87}}$ & $\mathbf{65.48_{\pm 0.76}}$   & $\mathbf{67.9_{\pm 1.45}}$ & $\mathbf{68.28_{\pm 0.18}}$ & $\mathbf{63.77_{\pm 1.37}}$   & $\mathbf{57.80_{\pm 2.38}}$ & $\mathbf{51.60_{\pm 0.41}}$ & $\mathbf{41.60_{\pm 3.94}}$  \\ \hline

GIN     & $48.93_{\pm 2.99}$ & $34.95_{\pm 0.86}$ & $14.53_{\pm 0.97}$  & $58.88_{\pm 0.57}$  & $53.80_{\pm 0.52}$ & $48.43_{\pm 0.69}$ & $39.25_{\pm 0.57}$  & $30.75_{\pm 1.45}$ & $22.35_{\pm 0.86}$\\ 

DisC$_{GIN}$     & $\mathbf{77.80_{\pm 1.33}}$ & $\mathbf{73.00_{\pm 0.61}}$ & $\mathbf{58.80_{\pm 1.66}}$  & $\mathbf{67.15_{\pm 0.79}}$  & $\mathbf{59.98_{\pm 0.62}}$ & $\mathbf{51.70_{\pm 0.34}}$  & $\mathbf{55.47_{\pm 0.98}}$  & $\mathbf{43.20{\pm 1.36}}$ & $\mathbf{31.33_{\pm 1.71}}$ \\ \hline

GCNII     & $53.50_{\pm 6.23}$ & $45.52_{\pm 2.26}$ & $32.6_{\pm 5.66}$  & $58.85_{\pm 1.89}$ & $53.98_{\pm 0.85}$ & $46.97_{\pm 1.38}$ & $39.93_{\pm 0.88}$ & $30.33_{\pm 1.17}$ & $23.09_{\pm 1.83}$ \\ 

DisC$_{GCNII}$     & $\mathbf{79.65_{\pm 2.13}}$  & $\mathbf{76.63_{\pm 1.38}}$ & $\mathbf{60.00_{\pm 5.66}}$   & $\mathbf{60.50_{\pm 2.77}}$& $\mathbf{63.05_{\pm 2.25}}$ & $\mathbf{61.78_{\pm 1.60}}$  & $\mathbf{56.23_{\pm 3.45}}$ & $\mathbf{49.10_{\pm 2.05}}$ & $\mathbf{41.05_{\pm 0.11}}$\\ \hline
\end{tabular}}
\end{table}

\begin{figure}[!htbp]
\centering
\subfigure[q]{
\includegraphics[ width=4cm]{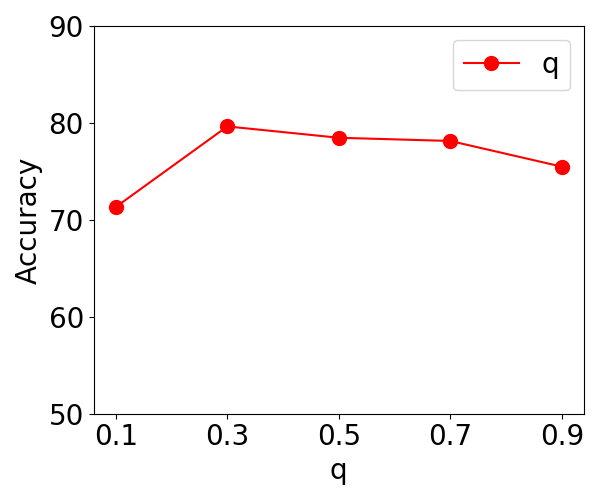}
\label{fig:q}
}
\subfigure[$\lambda_G$]{
\includegraphics[ width=4cm]{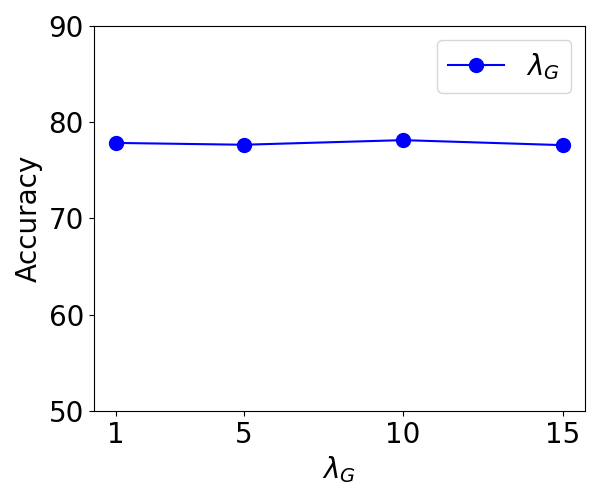}
\label{fig:G}
}
\caption{The hyperparameter experiments of $q$ and $\lambda_G$}
\label{fig:hyper}
\end{figure}

\textbf{Hyperparameter experiments} Fig.~\ref{fig:hyper} is the hyperparameter experiments of the degree of amplifying bias $q$ in GCE loss and the importance of generation component  $\lambda_G$.  For $q$, we fix $\lambda_G=10$ and vary $q$ from $\{0.1, 0.3, 0.5, 0.7, 0.9\}$. For $\lambda_G$, we fix $q=0.7$ and vary $\lambda_G$ from $\{1, 5, 10, 15\}$. From the results, we can see that our model achieves stable performance across different values of $q$ and $\lambda_G$. When $q=0.1$, it means the GCE loss will nearly reduce to normal CE loss. We can see the performance of $\text{DisC}_{GCN}$ is worse than other scenarios, demonstrating the effectiveness of utilizing GCE loss.
\subsection{Qualitative Evaluation}
\begin{figure*}[!htbp]
  \centering
  \includegraphics[width=12cm]{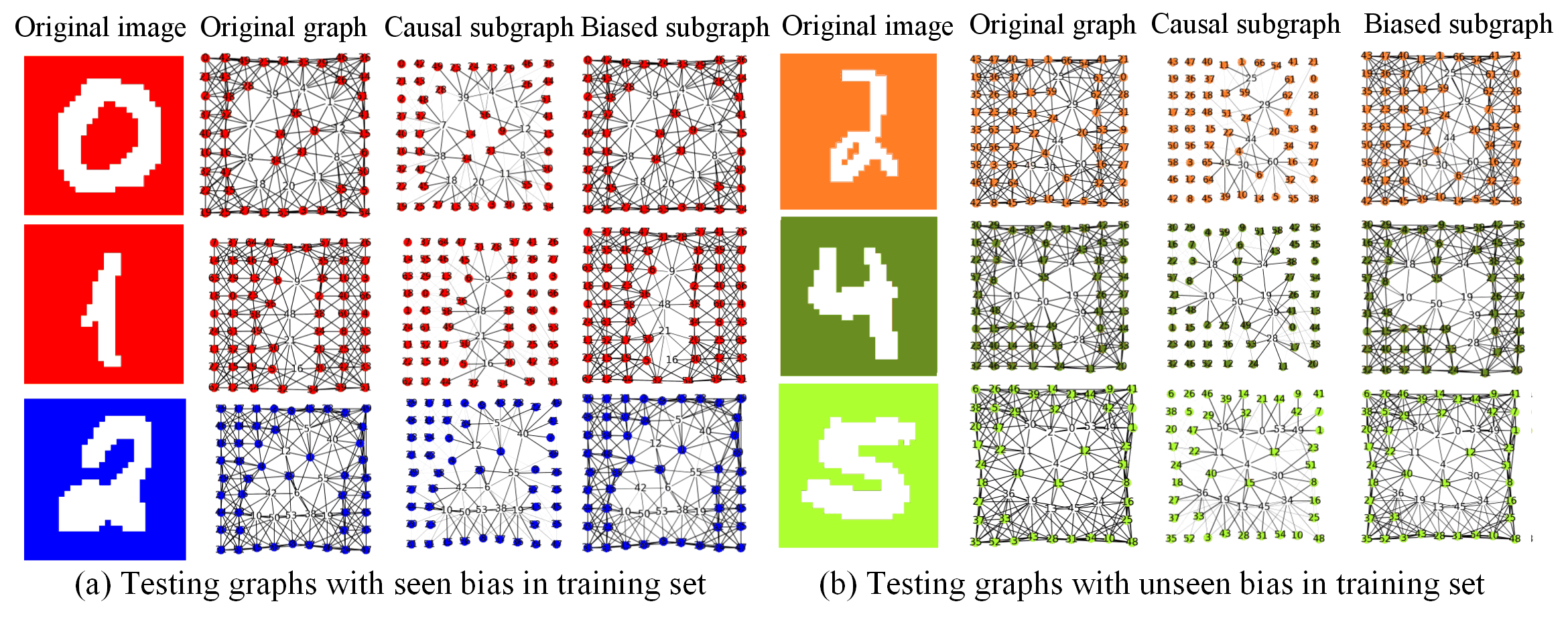}
  \caption{Visualization of subgraphs extracted by DisC. The width of edge is edge weight $c_{ij}$ or $b_{ij}$.}
  \label{fig::Mask}
\end{figure*}

\textbf{Visualization of edge mask.} To better illustrate the significant causal and bias subgraphs extracted by DisC$_{GCN}$, we visualize the original images, original graph, and corresponding causal subgraph and bias subgraph of CMNIST-75sp with 0.9 bias degree in Fig.~\ref{fig::Mask}, where the width of edge represents the value of learned weight $c_{ij}$ or $b_{ij}$. Fig.~\ref{fig::Mask}(a) shows the visualization results of testing graphs with the bias (color) that has been seen in the training set. As we can see, our model could discover the causal subgraphs where the most salient edges are in the digital subgraphs. With these causal subgraphs that highlight the structure information of digital, the GNNs will more easily extract this causal information. Fig.~\ref{fig::Mask}(b) shows the visualization results of testing graphs with unseen bias. According to the visualization, our model could still discover the causal subgraph outline, indicating our model could recognize causal subgraphs, whether the bias is seen or unseen. The visualization results of CFashion-75sp and CKuzushiji-75sp are shown in App.~\ref{appendix::experiments}.

% \usepackage{booktabs}
% \usepackage{booktabs}

\begin{figure}[!htbp]
\centering
\subfigure[$z_c$ labeled by digit.]{
\includegraphics[ width=3.4cm]{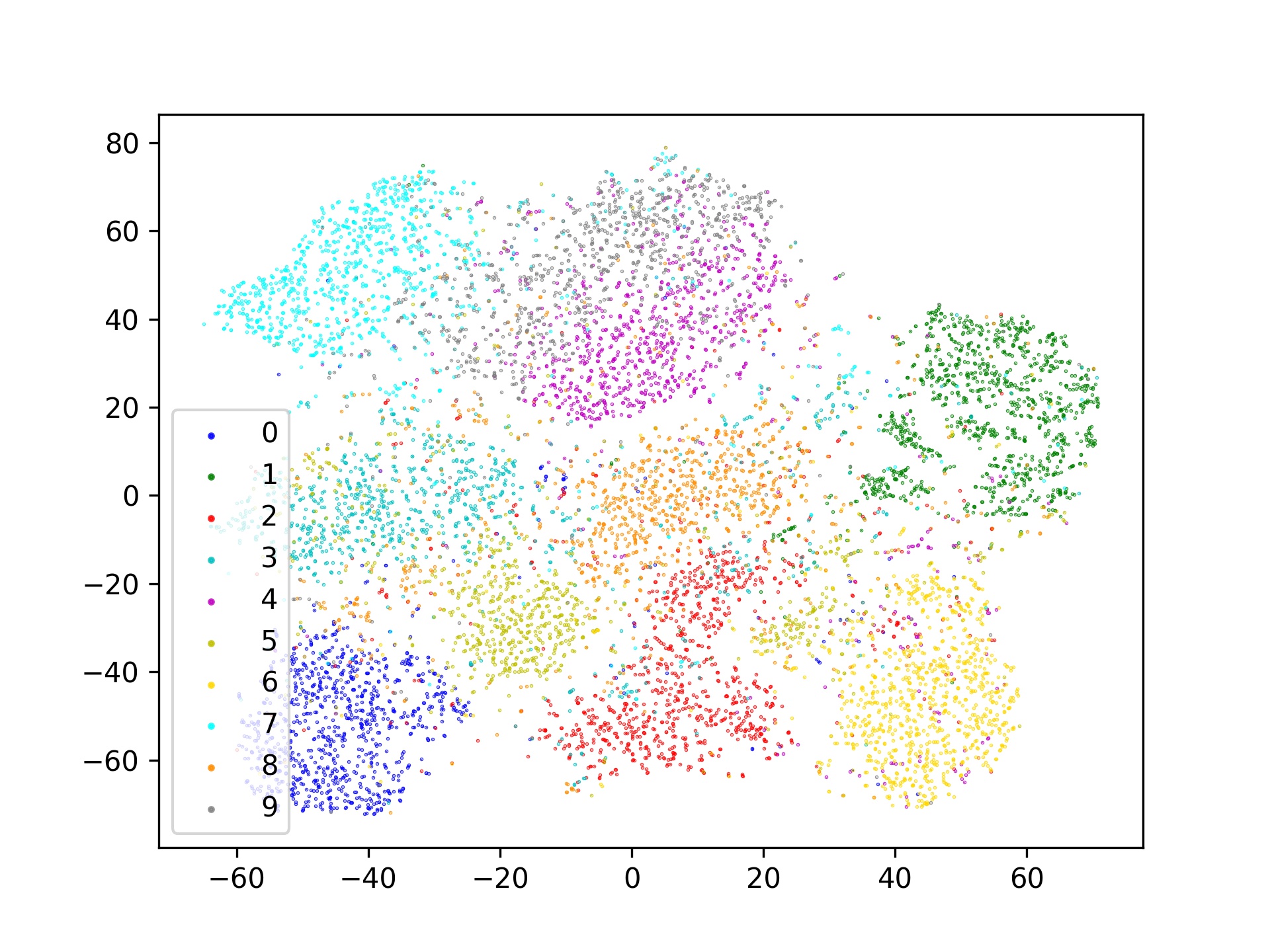}
\label{fig:3a}
}
\hspace{-10pt}
\subfigure[$z_c$ labeled by color.]{
\includegraphics[ width=3.4cm]{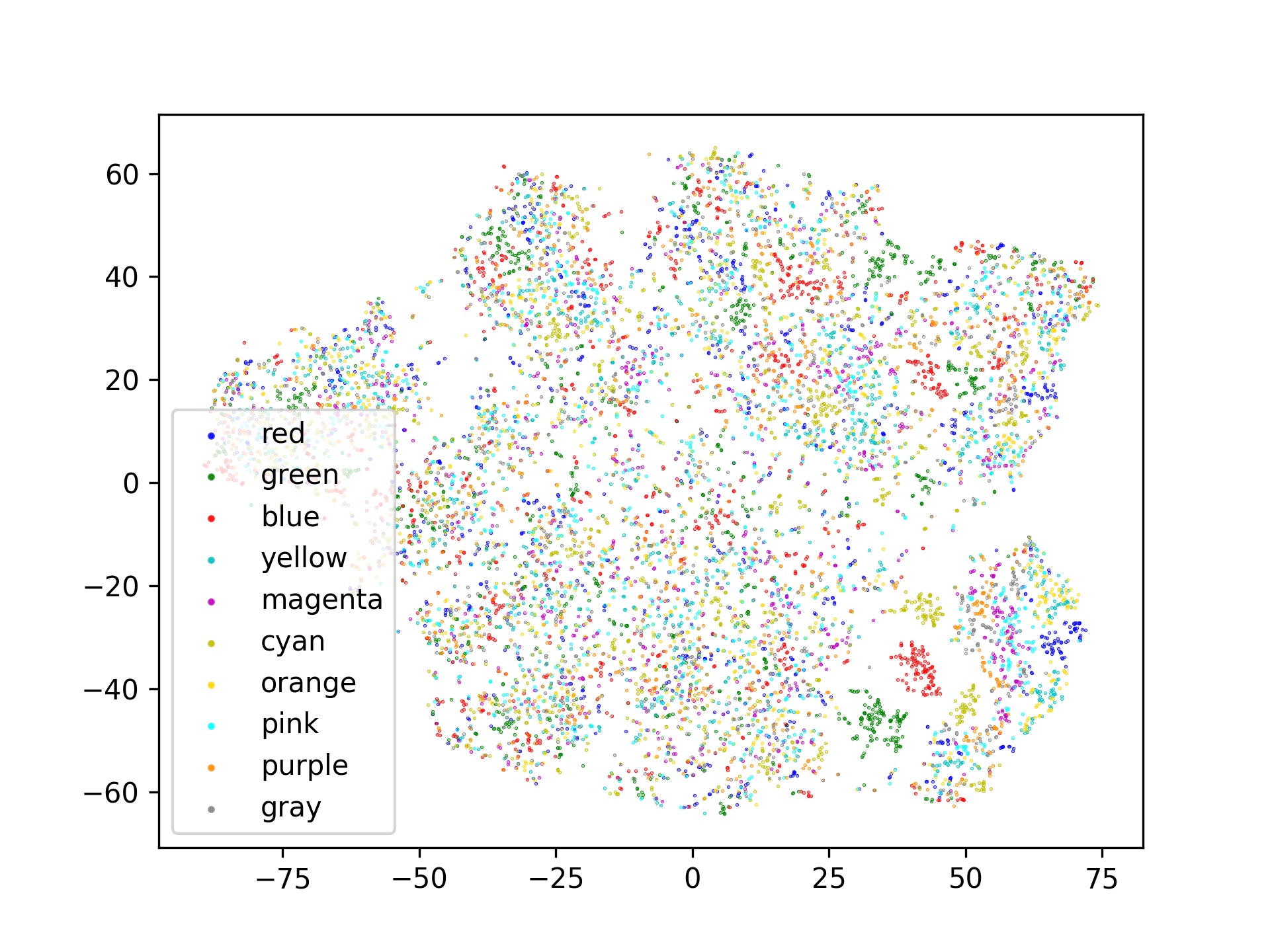}
\label{fig:3b}
}
\hspace{-10pt}
\subfigure[$z_b$ labeled by digit.]{
\includegraphics[ width=3.4cm]{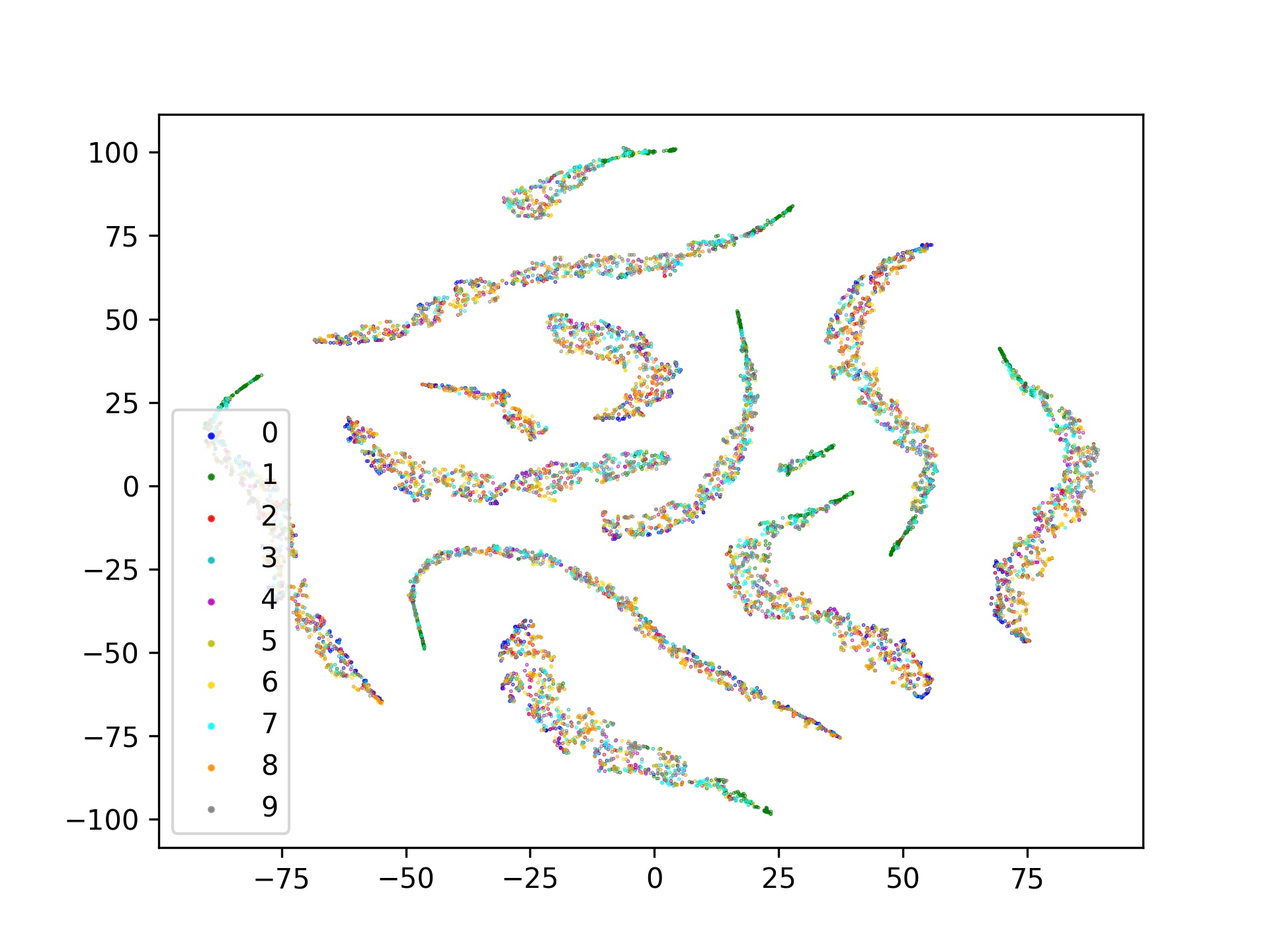}
\label{fig:3c}
}
\hspace{-10pt}
\subfigure[$z_b$ labeled by color.]{
\includegraphics[ width=3.4cm]{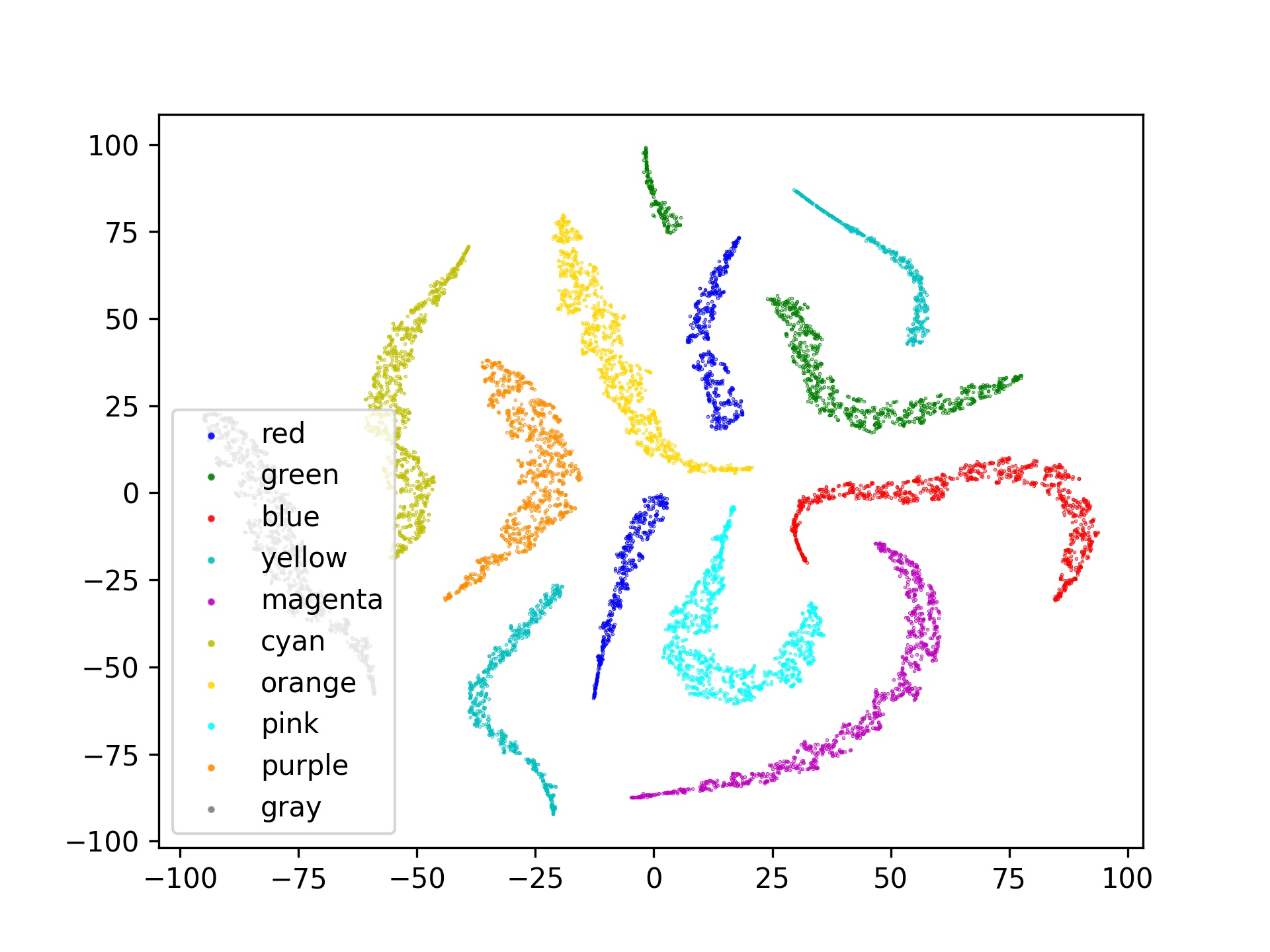}
\label{fig:3d}
}

\caption{Visualization of $z_c$ and $z_b$ with colors labeled by the digit and bias (color) labels. We observe that $z_c$ and $z_b$ are well clustered according to the groundtruth labels and bias labels, respectively.}
\label{fig:embed}
\end{figure}

\textbf{Projection of disentangled representation.} Fig.~\ref{fig:embed} shows the projection of latent vectors $z_c$ and $z_b$ extracted from the causal GNN $g_c$ and bias GNN $g_b$ of DisC$_{GCN}$, respectively, using t-SNE~\cite{t-sne} on CMNIST-75sp. Fig.~\ref{fig:embed} (a-b) are the projections of $z_c$ labeled by the target labels (digit) and bias labels (color), respectively. Fig.~\ref{fig:embed} (c-d) are the projections of $z_b$ labeled by the target labels and bias labels, respectively. We observe that $z_c$ are clustered according to the target labels while $z_b$ are clustered with the bias labels. And  $z_c$ are mixed with bias labels and $z_b$ are mixed with target labels. The results indicate that DisC successfully learns the disentangled causal and bias representations.

\begin{wrapfigure}{r}{4cm}

  \centering
  \includegraphics[width=4cm]{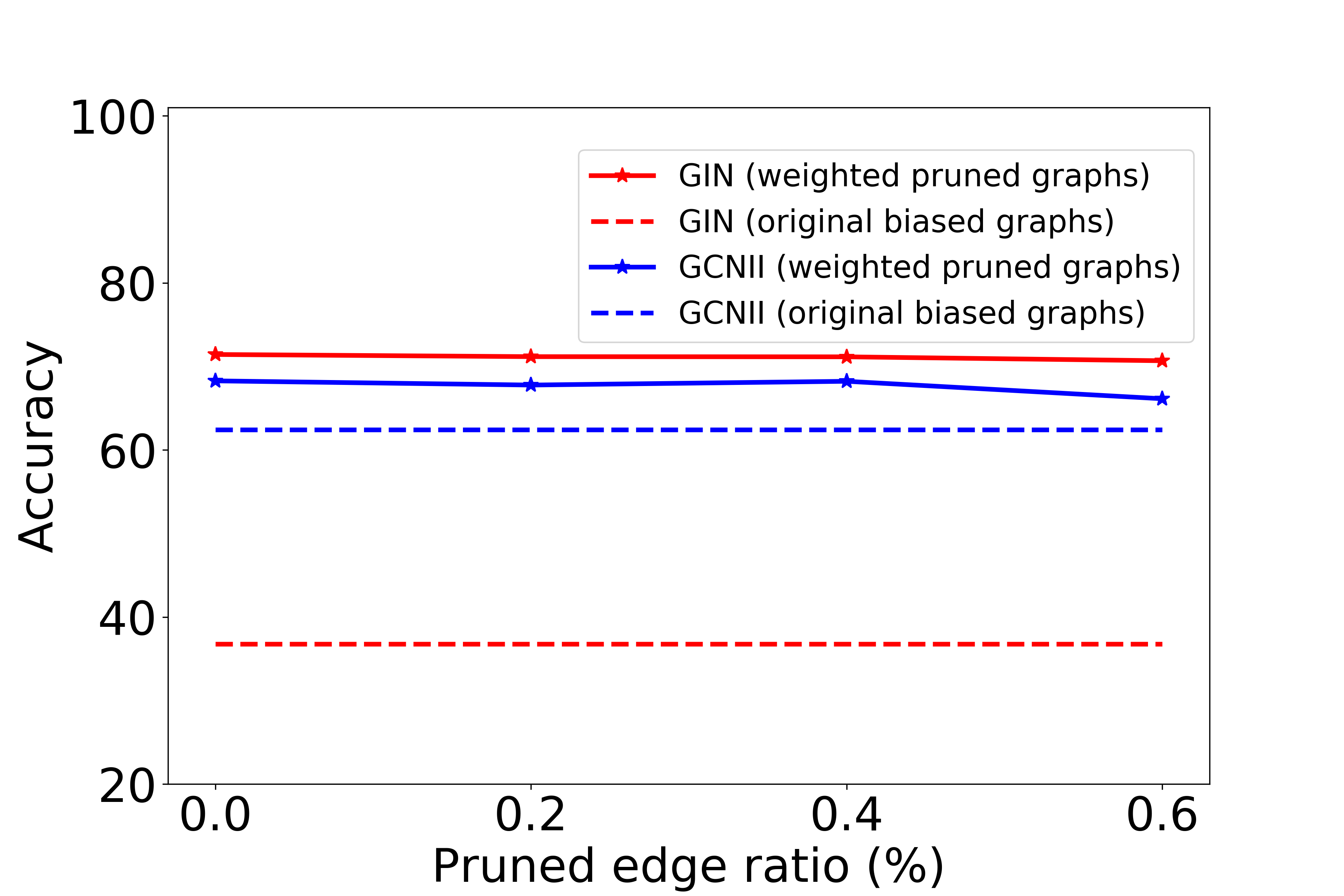}
  \caption{Performance of GIN and GCNII on the weighted pruned graphs found by DisC$_{GCN}$.}
  \label{fig::transfer}
\end{wrapfigure}
\textbf{Transferability of the learned mask.} As our model could extract GNN-independent subgraphs, the learning edge weights can be used to purify original biased graphs. These sparse subgraphs represent significant semantic information and can be universally transferred to any GNNs. To validate this point, we learn the edge mask by DisC$_{GCN}$ and prune the edges with least $\{0\%, 20\%, 40\%, 60\%\}$ weights while keeping the remaining edge weights. Then we train vanilla GIN and GCNII on these weighted pruned datasets. Fig.~\ref{fig::transfer} is the comparison of the results, where the dashed lines represent the results of base model on original biased graphs and the solid lines represent the performance of GNNs on weighted pruned datasets. The results show that the GNNs trained on the pruned datasets achieve better performances, indicating our learned edge mask has considerable transferability.

% \begin{figure}[!htbp]
% \centering
% \subfigure[GIN]{
% \includegraphics[ width=4cm]{Figures/GIN_sparisty.png}
% \label{fig:transfer_GIN}
% }
% \subfigure[GAT]{
% \includegraphics[ width=4cm]{Figures/GAT_sparisty.png}
% \label{fig:transfer_GAT}
% }
% \subfigure[GCNII]{
% \includegraphics[ width=4cm]{Figures/GCNII_sparisty.png}
% \label{fig:transfer_GCNII}
% }
% \caption{Performance of diverse GNNs on the sparse graphs found by DisC$_{GCN}$.}
% \label{fig:transfer}
% \end{figure}

\section{Conclusion}
\label{Sec::con}
In this paper, we are first to study the generalization problem of GNNs on severe bias datasets, which is crucial to study the transparently knowledge learning mechanism of GNNs. We analyze the problem in a causal view that the generalization of GNNs will be hindered by entangled representations as well as the correlation between causal and bias variables. To remove the impact from these two aspects, we propose a general disentangling framework, DisC, which extracts causal substructure and bias substructure by two different functional GNNs, respectively. After the representations are well-disentangled, we proliferate the counterfactual unbiased samples by randomly swapping the disentangled vectors. With the new constructed benchmarks, we clearly validate the effectiveness, robustness, interpretability, and transferability of our method.

%\textbf{Limitations and societal impacts.} Our method assumes that the graph consists of causal subgraph and bias subgraph. In reality, it may also exist non-informative subgraphs, which is neither causal nor biased for label. We would like to consider more fine-grained splitting of graphs in the future. When deploying GNNs to real-world applications, especially safety-critical fields, whether the results of GNNs are stable is an important factor. The demands for a stable model are universal and extensive such as in the field of disease prediction~\cite{sun2020disease}, traffic states prediction~\cite{cui2019traffic}, and financial applications~\cite{wang2021review}, where utilizing human-understandable causal knowledge for prediction is necessary.

\begin{ack}
This work is supported in part by the National Natural Science Foundation of China (No. U20B2045, 62192784, 62172052, 62002029, 62172052, U1936014). This work is also partially supported by the Natural Sciences and Engineering Research
Council (NSERC) Discovery Grant, the Canada CIFAR AI Chair Program, collaboration grants between Microsoft Research and Mila, Samsung Electronics Co., Ltd., Amazon Faculty Research Award,
Tencent AI Lab Rhino-Bird Gift Fund and a NRC Collaborative R\&D Project (AI4D-CORE-06).
This project was also partially funded by IVADO Fundamental Research Project grant PRF-2019-
3583139727. The work of Shaohua Fan is supported by the China Scholarship Council (No.202006470078). The computation resource of this project is supported by Compute Canada\footnote{https://www.computecanada.ca/}.
\end{ack}
%%% END INSTRUCTIONS %%%
\bibliographystyle{plain}
\bibliography{ref.bib}
\section*{Checklist}

\begin{enumerate}

\item For all authors...
\begin{enumerate}
  \item Do the main claims made in the abstract and introduction accurately reflect the paper's contributions and scope?
    \answerYes{}
  \item Did you describe the limitations of your work?
    \answerYes{See App. E.}
  \item Did you discuss any potential negative societal impacts of your work?
    \answerNo{We could not foresee any potential negative societal impacts of our work.}
  \item Have you read the ethics review guidelines and ensured that your paper conforms to them?
    \answerYes{}
\end{enumerate}

\item If you are including theoretical results...
\begin{enumerate}
  \item Did you state the full set of assumptions of all theoretical results?
    \answerYes{See Sec.~\ref{Sec:causal view}.}
        \item Did you include complete proofs of all theoretical results?
    \answerYes{}
\end{enumerate}

\item If you ran experiments...
\begin{enumerate}
  \item Did you include the code, data, and instructions needed to reproduce the main experimental results (either in the supplemental material or as a URL)?
    \answerYes{We provide the URL of code and data for reproducing the main results.}
  \item Did you specify all the training details (e.g., data splits, hyperparameters, how they were chosen)?
    \answerYes{See Sec.~\ref{Sec:Exp} and App. C.}
        \item Did you report error bars (e.g., with respect to the random seed after running experiments multiple times)?
    \answerYes{See Sec.~\ref{Sec:Exp}.}
        \item Did you include the total amount of compute and the type of resources used (e.g., type of GPUs, internal cluster, or cloud provider)?
    \answerYes{See Sec.~\ref{Sec:Exp}.}
\end{enumerate}

\item If you are using existing assets (e.g., code, data, models) or curating/releasing new assets...
\begin{enumerate}
  \item If your work uses existing assets, did you cite the creators?
    \answerYes{We construct new assets based on existing assets. We have cited them in Sec.~\ref{Sec:Exp}.}
  \item Did you mention the license of the assets?
    \answerYes{See App. C.}
  \item Did you include any new assets either in the supplemental material or as a URL?
    \answerYes{We release new assets through a URL in Sec.~\ref{Sec:Exp}.}
  \item Did you discuss whether and how consent was obtained from people whose data you're using/curating?
    \answerNo{The source data for generating our data is publicly available.}
  \item Did you discuss whether the data you are using/curating contains personally identifiable information or offensive content?
    \answerNo{We do not have personally identifiable information or offensive content.}
\end{enumerate}

\item If you used crowdsourcing or conducted research with human subjects...
\begin{enumerate}
  \item Did you include the full text of instructions given to participants and screenshots, if applicable?
    \answerNA{}
  \item Did you describe any potential participant risks, with links to Institutional Review Board (IRB) approvals, if applicable?
    \answerNA{}
  \item Did you include the estimated hourly wage paid to participants and the total amount spent on participant compensation?
    \answerNA{}
\end{enumerate}

\end{enumerate}
\clearpage
\appendix

\section{Preliminaries of Causal Inference}
\label{Sec::preli}
\subsection{Structural Causal Models}
In order to rigorously formalize our causal assumption behind the dataset, we resort to the Structural Causal Models, or SCM. SCM is a way of describing the relevant features (variables) of a particular problem and how they interact with each other. In particular, an SCM describes how the system assigns values to variables of interest.

\par Formally, an SCM consists of a set of \textit{exogenous variables} $U$ and a set of \textit{endogenous variables} $V$, and a set of functions $f$ that determines the values of variables in $V$ based on the other variables in the model. Casually, a variable $X$ is a \textit{direct cause} of a variables $Y$ if $X$ exists in the function that determines the value of $Y$. If $X$ is a direct cause of $Y$ or of any cause of $Y$, $X$ is a \textit{cause} of $Y$. Exogenous variables $U$ roughly means that they are external to the model, hence, in most scenarios, we choose not to explain how they are caused. Every endogenous variable is a descendant of at least one exogenous variable. Exogenous variables can only be the root variables. If we know the value of every exogenous variable, with the functions in $f$ , we can perfectly determine the value of every endogenous variable. In many cases, we usually assume that all exogenous variables are unobserved variables like noise and are independently distributed with an expected value zero, so we only interest with the interaction with endogenous variables. Every SCM is associated with a \textit{graphical causal model} or simply referred to ``casual graph''. Causal graph consists of nodes representing the variables in $U$ and $V$, and the direct edges between the nodes representing the functions in $f$. Note the in our SCM in Section 3.2, we only show the endogenous variables we are interested in.

\subsection{$d$-separation/connection}
\begin{figure*}[!htbp]
  \centering
  \includegraphics[width=14cm]{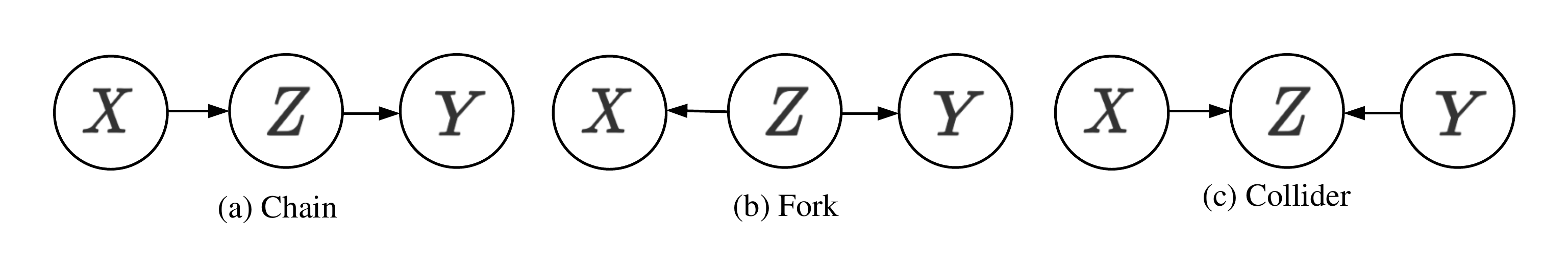}
  \caption{Three basic relations in causal graph. }
  \label{fig::causal}
\end{figure*}

Given an SCM, we are particularly interested in (conditional) dependence information that is embedded in the model. There are three basic relationships of variables in an SCM, $i.e.,$ chains, forks and colliders, as shown in Fig.~\ref{fig::causal}.  For chains and forks, $X$ and $Y$ would be dependent if $Z$ is not in the conditional set, $i.e.,$ the path is unblocked, and vice versa. And for colliders, $X$ and $Y$ would be independent if $Z$ is not in the conditional set, $i.e.,$ the path is blocked. Built upon these rules, $d$-separation is a criterion that can be applied in causal graphs of any complexity in order to predict dependencies that are shared by all datasets generated by the graph~\cite{glymour2016causal}.  Two nodes $X$ and $Y$ are $d$-separated if every path between them is blocked. If even one path between $X$ and $Y$ is unblocked, $X$ and $Y$ are $d$-connected. Formally, we have following definition of $d$-separation:

\begin{myDef}[$d$-separation~\cite{glymour2016causal}] A path $p$ is blocked by a set of nodes $Z$ if and only if 

1. p contains a chain of nodes $A\rightarrow B\rightarrow C$ or a fork $A\leftarrow B \rightarrow C$ such that the middle node $B$ is in $Z$ ($i.e.,$ $B$ is conditioned on), or

2. $p$ contains a collider $A\rightarrow B \leftarrow C$ such that the collision node $B$ is not in $Z$, and no descendant of $B$ is in $Z$.
\end{myDef}

With this principle, we could find that the paths \textbf{(1) $\mathbf{B\rightarrow G \rightarrow E \rightarrow Y}$} and  \textbf{(2)  $\mathbf{B\leftrightarrow C \rightarrow Y}$} in Section 3.2 are unblocked paths, which would induce unexpected correlation between bias variable $B$ and prediction $Y$.

More knowledge of causal inference, please refer to~\cite{glymour2016causal, pearl2009causality}.

\section{Algorithm}
\label{sec::alg}
\begin{algorithm}[H]
    \caption{Disentangled Casual Substructure Learning}
\label{alg1}
		%     \SetAlgoNoLine % 不要算法中的竖线
\SetKwInOut{Input}{\textbf{Input}}\SetKwInOut{Output}{\textbf{Output}}\SetKwInOut{Initialization}{\textbf{Initialization}} % 替换关键词
\Input{graph $G$, label $y$, max iteration $T$, generation iteration $t_{gen}$}
		    \Output{Learned edge mask generator MLP, two GNN networks $(g_c, g_b)$, and two classifiers $(C_c, C_b)$
		      }
		    \Initialization{iteration $t=0$; Initialize MLP, $(g_c, g_b)$ and  $(C_c, C_b)$}
\BlankLine %\varphi_n\bm{W}\vec{y}_n
    \While{not converged \rm{or} $t<T$}
    {
     Extract subgraphs $G_c$ and $G_b$ based on edge mask generator; \\
     Encode subgraphs $G_c$ and $G_b$ into $z_c$ and $z_b$ via $g_c(G_c)$ and $g_b(G_b)$;\\
     Concatenate $z = [z_c; z_b]$; \\
     Update  $(g_c, g_b)$ and  $(C_c, C_b)$ via $L_D$ in Eq. (6);\\
     if $t>t_{gen}$:\\
      \quad Randomly swap $z=[z_c; z_b]$ into $z_{unbiased}=[z_c; \hat{z_b}]$; \\
      \quad Update  $(g_c, g_b)$ and  $(C_c, C_b)$ via  $L$ in Eq. (8); \\
      $t = t+1$;
    }
\end{algorithm}

\section{Experimental Details}
\label{Append::Experimental Details}
\subsection{Datasets details}
\label{appendix datasets}
\setlength\tabcolsep{2pt}
\begin{table*}[ht]
  \caption{Statistics of Biased Graph Classification Datasets.}
  \label{sample-table}
  \centering
  \resizebox{1\textwidth}{!}{
  \begin{tabular}{llllllllll}
    \toprule
    \textbf{Dataset}   & \textbf{Causal subgraph type} & \textbf{Bias subgraph type} & \textbf{\#Graphs(train/val/test)}  & \textbf{\#Classes} & \textbf{\#Avg. Nodes}  & \textbf{\#Avg. Edges} & \textbf{Node feat (dim.)}  &\textbf{Bias degree} & \textbf{Difficulties}\\
    \midrule
    CMNIST-75sp    & Digit subgraph & Color background subgraph   & 10K/5K/10K  &  10 & 61.09 & 488.78 & Pixel+Coord (5)  & 0.8/0.9/0.95 & Easy\\
    CFashion-75sp   & Fashion product subgraph & Color background subgraph    &10K/5K/10K  & 10 & 61.03 & 488.26 & Pixel+Coord (5) & 0.8/0.9/0.95 & Medium\\
    CKuzushiji-75sp & Hiragana subgraph & Color background subgraph & 10K/5K/10K & 10 & 52.87 & 423.0 & Pixel+Coord (5) & 0.8/0.9/0.95 & Hard \\
    \bottomrule
  \end{tabular}}
  \label{tab::data_stats}
\end{table*}
We summarize statistics of datasets constructed in this paper in Table~\ref{tab::data_stats}. Note that the bias degree of validation set is 0.5, we use it to adjust the learning rate during training process. Without loss of any generality, here we subsample original 60K training samples into 10K training samples to make the training process more efficient. One could easily construct full dataset with our method. Each graph of CFashion-75sp is labeled by the category of fashion product it belongs to and each graph of CKuzushiji-75sp is labeled by one of 10 Hiragana characters. Moreover, we would like to list the map between label and predefined correlated color for all datasets in Table~\ref{tab::map}. The links for source image datasets are as follows:
\begin{enumerate}
    \item MNIST: http://yann.lecun.com/exdb/mnist/.
    \item Fashion-MNIST: https://github.com/zalandoresearch/fashion-mnist. MIT License.
    \item Kuzushiji-MNIST: https://github.com/rois-codh/kmnist. CC BY-SA 4.0 License.
\end{enumerate}

\begin{table*}[ht]
  \caption{Mapping between label and color.}
  \centering
  
  \begin{tabular}{ll ll}
    \toprule
    \textbf{Label}   & \textbf{Color (RGB)} &\textbf{Label}   & \textbf{Color (RGB)}\\
    \midrule
    0 & (255, 0, 0) & 5 & (0, 255, 255) \\
    1 & (0, 255, 0) & 6 & (255, 128, 0)\\
    2 & (0, 0, 255) & 7 & (255, 0, 128)\\
    3 & (225, 225, 0) & 8 & (128, 0, 255)\\
    4 & (225, 0, 225) & 9 & (128, 128, 128)\\
    \bottomrule
  \end{tabular}
  \label{tab::map}
\end{table*}
For unbiased testing dataset with unseen bias used in Table~\ref{tab:unseen}, the RGB value of predefined color set is \{(199, 21, 133), (255, 140, 105), (255, 127, 36), (139, 71, 38),  (107, 142, 35), (173, 255, 47), (60, 179, 113), (0, 255, 255), (64, 224, 208), (0, 191, 255)\}.

\subsection{Experimental setup} 
\label{appendex setup}
For GCN and GIN, we use the same model architectures as~\cite{hendrycks2019benchmarking}\footnote{https://github.com/graphdeeplearning/benchmarking-gnns.}, which have 4 layers, and 146 hidden dimension for GCN and 110 hidden dimension for GIN. And GIN utilizes its GIN0 variant. For GCNII, it has 4 layers and 146 hidden dimension. DIR\footnote{https://github.com/Wuyxin/DIR-GNN.} utilizes the default parameters in original paper for MNIST-75sp dataset. For causal GNN or bias GNN in our model, it has the same architecture with base model. We optimize all models with the Adam~\cite{kingma2014adam} optimizer and 0.01 learning rate with for all experiments. The batch-size for all the methods is 256. We train all the models with 200 epochs and set the generation iteration of our method $t_{gen}$ as 100. For our model, we set $q$ of GCE loss as $0.7$ and $\lambda_G$ as 10 for all experiments. Our substructure generator is a two-layer MLP, whose activation function is sigmoid function. For StableGNN, we use their GraphSAGE variant. For other baselines, we use their default hyperparameters. LDD\footnote{https://github.com/kakaoenterprise/Learning-Debiased-Disentangled} has same hyparameters with our model. To better reflect the performance of unbiased sample generation, we take the performances of last step as final results.  All the experiments are conducted on the single NVIDIA V100 GPU.
\section{Visualization of CFashion-75sp and CKuzushiji-75sp}
\label{appendix::experiments}
\begin{figure*}[!htbp]
  \centering
  \includegraphics[width=12cm]{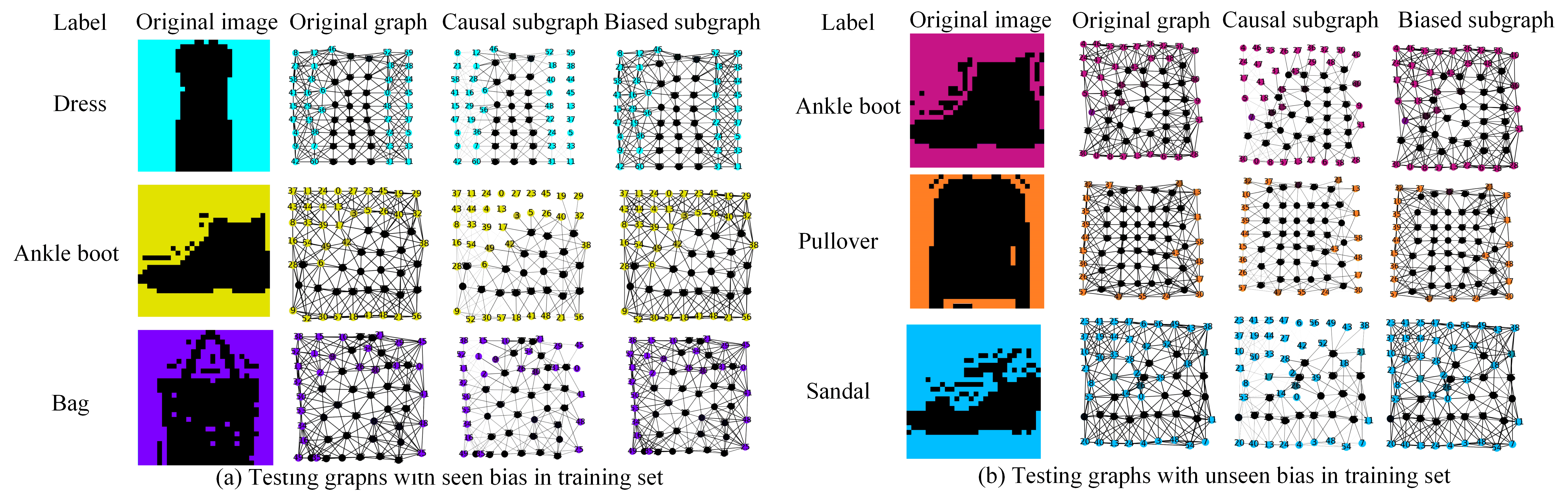}
  \caption{Visualization of subgraphs extracted by the mask generator from CFashion-75sp. }
  \label{fig::Mask2}
\end{figure*}

\begin{figure*}[!htbp]
  \centering
  \includegraphics[width=12cm]{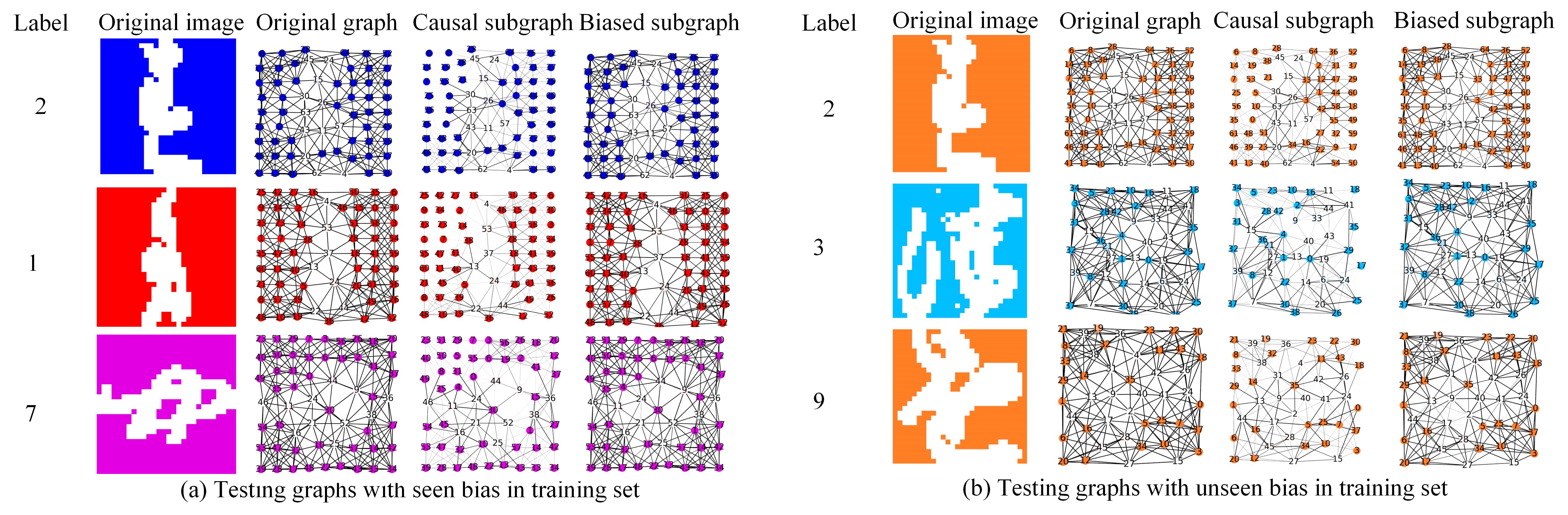}
  \caption{Visualization of subgraphs extracted by the mask generator from CKuzushiji-75sp.} 
  \label{fig::Mask3}
\end{figure*}

Figure~\ref{fig::Mask2} and Figure~\ref{fig::Mask3} are visualization results of  CFashion-75sp and  CKuzushiji-75sp dataset. As we can see, our model could also discover reasonable causal subgraphs for these challenging datasets.

% \subsection{Hyperparameter experiments}
% Fig.~\ref{fig:hyper} is the hyperparameter experiments of the degree of amplifying bias $q$ in GCE loss and the importance of generation component  $\lambda_G$.  For $q$, we fix $\lambda_G=10$ and vary $q$ from $\{0.3, 0.5, 0.7, 0.9\}$. For $\lambda_G$, we fix $q=0.7$ and vary $\lambda_G$ from $\{1, 5, 10, 15\}$. From the results, we can see that our model achieves stable performance across different values of $q$ and $\lambda_G$. When $q=0.1$, it means the GCE loss will nearly reduce to normal CE loss. We can see the performance of $\text{DisC}_{GCN}$ is worse than other scenarios, demonstrating the effectiveness of utilizing GCE loss.
% \begin{figure}[!htbp]
% \centering
% \subfigure[q]{
% \includegraphics[ width=5cm]{Figures/q.png}
% \label{fig:illustration}
% }
% \subfigure[$\lambda_G$.]{
% \includegraphics[ width=5cm]{Figures/G.png}
% \label{fig:motivating results}
% }
% \caption{The hyperparameter experiments of $q$ and $\lambda_G$}
% \label{fig:hyper}
% \end{figure}

\section{Limitations and societal impacts}
Our method assumes that the graph consists of causal subgraph and bias subgraph. In reality, it may also exist non-informative subgraphs, which is neither causal nor biased for label. We would like to consider more fine-grained splitting of graphs in the future. When deploying GNNs to real-world applications, especially safety-critical fields, whether the results of GNNs are stable is an important factor. The demands for a stable model are universal and extensive such as in the field of disease prediction~\cite{sun2020disease}, traffic states prediction~\cite{cui2019traffic}, and financial applications~\cite{wang2021review}, where utilizing human-understandable causal knowledge for prediction is necessary.

\end{document}